\documentclass{article}

\usepackage[preprint]{neurips_2023}

\usepackage[utf8]{inputenc} %
\usepackage[T1]{fontenc}    %
\usepackage{xcolor}
\usepackage{hyperref}       %
\definecolor{mydarkblue}{rgb}{0,0.08,0.45}
\hypersetup{
colorlinks=true,
citecolor=mydarkblue,
filecolor=mydarkblue,
urlcolor=mydarkblue,
}
\usepackage{url}            %
\usepackage{booktabs}       %
\usepackage{amsfonts}       %
\usepackage{amsmath}        %
\usepackage{amssymb}        %
\usepackage{nicefrac}       %
\usepackage{microtype}      %
\usepackage{xcolor}         %
\usepackage{multirow}
\usepackage{svg}
\usepackage{longtable}
\usepackage{dcolumn}
\usepackage{chngpage}
\usepackage{textcomp}
\newcolumntype{d}[1]{D{.}{.}{#1}}

\newcommand{\usmvtwo}{USM-v2 }

\DeclareUnicodeCharacter{2212}{-}  %

\usepackage{xspace}
\newcommand{\ttron}{Translatotron~2\xspace}
\newcommand{\ours}{AudioPaLM\xspace}

\title{AudioPaLM: A Large Language Model That Can Speak and Listen}

\author{
\and Paul K.~Rubenstein\thanks{Authors have contributed equally to this work.}    
\and Chulayuth Asawaroengchai\footnotemark[1]
\and Duc Dung Nguyen\footnotemark[1]
\and Ankur Bapna 
\and Zalán Borsos
\and Félix de Chaumont Quitry  
\and Peter Chen
\and Dalia El Badawy
\and Wei Han
\and Eugene Kharitonov
\and Hannah Muckenhirn
\and Dirk Padfield
\and James Qin
\and Danny Rozenberg
\and Tara Sainath
\and Johan Schalkwyk
\and Matt Sharifi
\and Michelle Tadmor Ramanovich
\and Marco Tagliasacchi
\and Alexandru Tudor
\and Mihajlo Velimirović
\and Damien Vincent
\and Jiahui Yu
\and Yongqiang Wang
\and Vicky Zayats
\and Neil Zeghidour
\and Yu Zhang
\and Zhishuai Zhang
\and Lukas Zilka
\and Christian Frank
}

\date{\today}

\begin{document}

\maketitle
\begin{center}
  \vspace{-1.0cm}
  Google
  \vspace{0.5cm}
\end{center}

\begin{abstract}
We introduce \ours{}, a large language model for speech understanding and generation.
\ours{} fuses text-based and speech-based language models, PaLM-2~\citep{palmv2} and AudioLM~\citep{borsos2022audiolm}, into a unified multimodal architecture that can process and generate text and speech with applications including speech recognition and speech-to-speech translation. 
\ours{} inherits the capability to preserve paralinguistic information such as speaker identity and intonation from AudioLM and the linguistic knowledge present only in text large language models such as PaLM-2.
We demonstrate that initializing \ours{} with the weights of a text-only large language model improves speech processing, successfully leveraging the larger quantity of text training data used in pretraining to assist with the speech tasks. The resulting model significantly outperforms existing systems for speech translation tasks and has the ability to perform zero-shot speech-to-text translation for many languages for which input/target language combinations were not seen in training. \ours{} also demonstrates features of audio language models, such as transferring a voice across languages based on a short spoken prompt. We release examples of our method at: \href{https://google-research.github.io/seanet/audiopalm/examples}{\texttt https://google-research.github.io/seanet/audiopalm/examples}.

\end{abstract}

\section{Introduction}

Large language models (LLMs) \citep{gpt3, gopher, chowdhery2022palm} excel at generating text for tasks that require the modeling of complex interactions as well as knowledge retrieval, such as open-domain question answering or few-shot machine translation \citep{palmv2}. The remarkable generative abilities of the underlying system — a Transformer \citep{attentionvaswani} trained to predict sequences of discrete tokens — have been subsequently extended to continuous, natural signals with images \citep{parti} or audio waveforms \citep{gslm, audiogen, valle} being converted into a stream of discrete units through a lossy compression algorithm and then modeled in a sequential fashion as would be text.

In the context of audio generation, the AudioLM framework \citep{borsos2022audiolm} has introduced a hierarchical approach which combines two types of audio tokens, with high-level coarse tokens extracted from self-supervised embeddings \citep{w2vbert} being used to condition the generation of lower-level codes of a neural codec \citep{zeghidour2021soundstream}. This general framework, which makes little assumptions about the nature of the modeled audio signals, has been used to generate speech and music \citep{kharitonov2023speartts,agostinelli2023musiclm,singsong}. 
In the particular case of text-to-music \citep{agostinelli2023musiclm} or text-to-speech \citep{kharitonov2023speartts}, a Transformer model takes text tokens as inputs and generates audio tokens, such that text and audio vocabularies do not interact with each other. Such models could naturally be converted into, respectively, music captioning and speech recognition systems by swapping their inputs and outputs. Following this observation, combining text and audio vocabularies into a multimodal, single vocabulary would allow for training a single model in both directions.

\looseness=-1
In this work, we introduce AudioPaLM, a multimodal generative model of speech and text. At the heart of AudioPaLM is a joint vocabulary that can represent speech and text with a limited number of discrete tokens which, combined with an elementary markup description of tasks, allows training a single decoder-only model on a mixture of tasks that involve arbitrarily interleaved speech and text. 
This includes speech recognition, text-to-speech synthesis, and speech-to-speech translation, unifying tasks that are traditionally 
solved by heterogeneous models into a single architecture and training run. Moreover, as the underlying architecture of AudioPaLM is a large Transformer model, we can initialize its weights with those of a large language model pretrained on text which allow it to benefit from the linguistic and common sense knowledge of models such as PaLM~\citep{chowdhery2022palm} or PaLM~2~\citep{palmv2}. In particular, we show in Section ~\ref{subsec:exp-palm-2} how the model's translation capability is derived from the translation capability of the underlying text model.
The contributions of this work are:

\begin{itemize}
    \item We present a unified speech-text LLM, capable of consuming and producing both speech and text, and leveraging the existing capabilities of PaLM~\cite{chowdhery2022palm} and PaLM-2~\citep{palmv2} coming from text-only pretraining.
    \item This unified approach across modalities allows training \ours{} on a mixture of tasks such as Automatic Speech Recognition (ASR), Automatic Speech Translation (AST) and Speech-to-Speech Translation (S2ST), achieving state of the art results on AST and S2ST benchmarks, and competitive performance on ASR benchmarks.
    \item Leveraging AudioLM's audio prompting~\citep{borsos2022audiolm}, our model performs S2ST with voice transfer of unseen speakers, surpassing existing methods in terms of speech quality and voice preservation, as measured by both objective and subjective evaluations.
    \item Our model exhibits zero-shot capabilities, performing AST with speech input/target language combinations that were not seen in training.
\end{itemize}

The remainder of this paper is organized as follows: in Section~\ref{sec:related-work} we discuss the relation to existing work. In Section~\ref{sec:method} we describe our method.
In Section~\ref{sec:data-etc} we provide details about the data we use, and other technical details as a prelude to the experiments.
In Section~\ref{sec:experiments} we present our experimental results including a series of ablations to determine the influence of various design choices. 
We conclude in Section \ref{sec:conclusion}.

\section{Related work \label{sec:related-work}}

\subsection{Multimodal fusion}

\looseness=-1
Encoder-based models are used to learn features which can be used for downstream tasks. 
By learning joint representations of both modalities together, the goal is that in addition to the learned features being richer than they would be with each modality treated separately, they are aligned with one another, improving their performance when used for inter-modality tasks.
Such approaches have been applied in audio \citep{chen2022maestro, bapna2022mslam, zhang2023usm} and in vision \citep{chen2020uniter, gan2020large, fu2021violet} as well as combining both audio and video inputs~\citep{shi2022audiovisual}.

Similar to BERT \citep{devlin2018bert}, such encoders may be trained with a masked language model objective for both the multimodal setting as in previously mentioned works and for the unimodal setting \citep{baevski2020wav2vec, hsu2021hubert, chiu2022bestrq}.
They may alternatively be trained in a contrastive manner \citep{radford2021clip, yuan2021florence, yu2022coca} resulting in separate encoders for each modality with each informed by the other due to the contrastive objective.

A line of work on multimodal encoder-decoder models (also known as \emph{Vision Language Models} in the vision literature) has sought to fuse text-decoders with advances in non-text encoder models. Examples include Flamingo \citep{alayrac2022flamingo} and PaLI \citep{chen2022pali} in the vision domain, and Whisper \citep{radford2022whisper} in the audio domain.
The general idea of these approaches is to take an audio or vision encoder and a text decoder and to combine them, either with adapter layers as in Flamingo and Whisper, or by merging via a separate encoder as in PaLI.

Both PaLI and Flamingo use pretrained components.
The advantage of this is that individual components can be frozen while finetuning the model on multimodal data (Whisper does not use a pretrained encoder or decoder and so does not freeze individual components).
The disadvantage is that such models are constrained to only output text, since the decoder is text-only. 
In contrast, our proposed approach results in a decoder-only model which models sequences of arbitrary audio and text tokens. 
This is similar to the approach taken by \cite{wang2022ofa} except that we use a single decoder-only model and all audio seen by the model is tokenized, whereas \cite{wang2022ofa} use an encoder-decoder architecture and use continuous inputs and tokenized outputs for images.

\subsection{Generating audio with language models}
Recent work~\citep{gslm, valle} has explored generating speech by modeling discretized representations as target tokens of an autoregressive Transformer~\citep{attentionvaswani} network. Such discrete tokens can be extracted from self-supervised speech representations~\citep{oord2018cpc,baevski2020wav2vec,hsu2021hubert,w2vbert}, modeling long-term patterns in audio sequences while providing limited reconstruction quality, or from a neural codec~\citep{zeghidour2021soundstream,encodec}, providing high-fidelity reconstruction but with less temporal structure. AudioLM ~\citep{borsos2022audiolm} addresses this dichotomy by introducing a hierarchical approach, where a first stage produces ``semantic'' tokens from a self-supervised w2v-BERT system~\citep{w2vbert}, which a second stage then uses as conditioning to generate the ``acoustic'' tokens of a SoundStream~\citep{zeghidour2021soundstream} neural codec. This joint modeling of semantic and acoustic tokens allows the model to learn linguistic structure from the syntactic to the lexical and phonetic levels from speech-only corpora, without any textual guidance, while generating realistic speech from arbitrary speakers and in diverse acoustic conditions.

SPEAR-TTS~\citep{kharitonov2023speartts} combines the decoder-only generator of AudioLM with a text encoder, such that the model can perform text-to-speech synthesis. By leveraging pretraining and backtranslation~\citep{Sennrich2016}, SPEAR-TTS can be trained with only 15 minutes of labeled speech. The ability of this model to learn a mapping between text and semantic tokens in such a low-data regime suggests that these representations are very close, yet the model's encoder-decoder architecture specifically ingests text and outputs audio, such that both vocabulary of tokens (text and semantic) are disjoint and modeled separately. SpeechLM~\citep{Hassid2023TextuallyPS} also exploits the similarity between text and semantic tokens by initializing a decoder-only audio generator with the weights of a pretrained text-based language model. While this allows some transfer of knowledge from text-to-speech modeling, the resulting architecture is not multimodal: semantic tokens replace the text vocabulary ---rather than extending it--- and the model is finetuned on speech-only data. AudioPaLM bridges these gaps and combines semantic tokens and text into an extended, multimodal set of tokens used interchangeably as inputs and outputs, such that text-only language model pretraining can be used to initialize a decoder-only model that can then be finetuned on a mixture of tasks that map freely between speech and text (e.g. speech-to-text, text-to-speech or speech-to-speech).

\subsection{Speech-to-speech translation}

The field of speech-to-speech translation (S2ST) focuses on converting spoken language from one language to another, facilitating communication between individuals who speak different languages. Conventional automatic speech-to-speech translation systems are typically composed of a cascade of three components: automatic speech recognition (ASR), text-to-text machine translation (MT), and text-to-speech (TTS) synthesis  \citep{lavie1997janus, wahlster2000verbmobil, nakamura2006atr}.  However, these cascade-based approaches primarily focus on the text and may overlook important aspects such as para-linguistic features, computational efficiency, compound errors, and the accurate handling of proper names, nouns, and non-verbal communication that do not require translation.

Direct speech-to-speech translation systems \citep{jia2019direct,kano2021transformer,jia2022translatotron} are trained end-to-end operating on the audio spectrogram domain without relying on text representation at inference time. In these systems, the synthesized audio has access to acoustic information in source speech and can potentially learn to preserve acoustic features and reduce compound errors and computational requirements. 

There are other cascaded S2ST systems that utilize learned discrete speech representations as an intermediate representation \citep{tjandra2019speech,zhang2020uwspeech,lee2021direct,ma2021direct,lee2021textless}. In these systems the translation operates in learned discrete representation space allowing to learn alignment in the discrete domain, and simplify leveraging of text pre-training.
Lastly, there are other S2ST approaches that improve on performance, efficiency, and data requirements. \citet{jia2022leveraging} and \citet{wei2022joint} leveraged weakly supervised data and component pre-training to improve translation accuracy while requiring little parallel speech data.

\section{Method\label{sec:method}}

\begin{figure}
    \centering
    \includegraphics[width=\columnwidth]{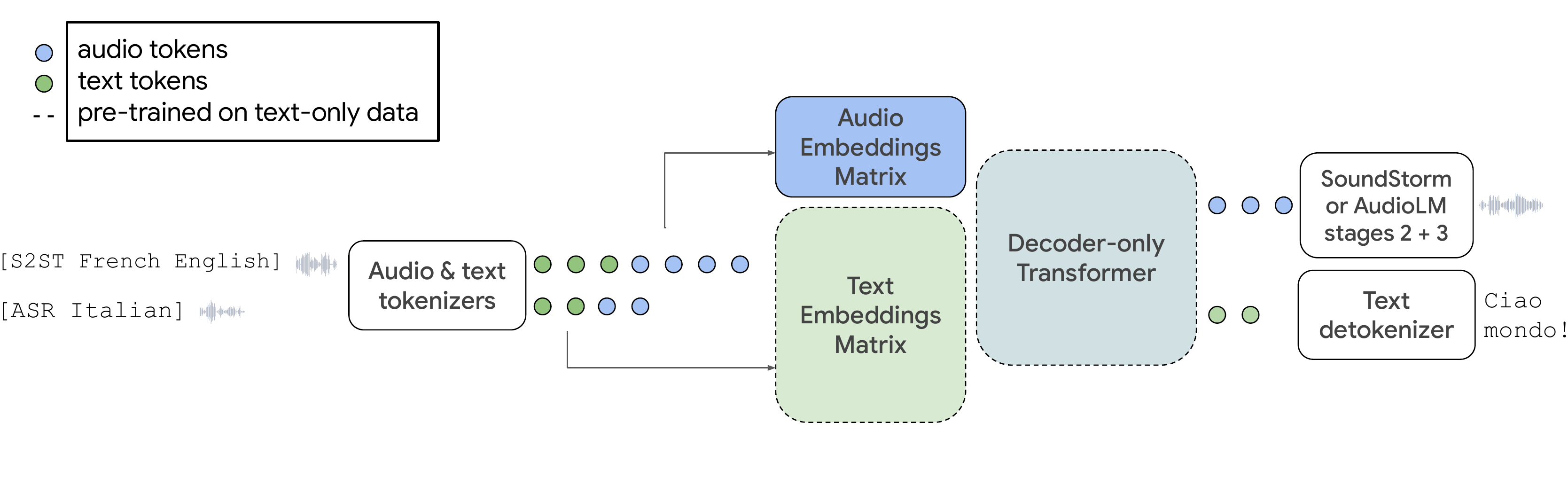}
    \caption{The AudioPaLM model, illustrated on speech-to-speech translation and automatic speech recognition. We take a pretrained text-only model (dashed lines) and expand its embeddings matrix to model a new set of audio tokens.
    The model architecture is otherwise unchanged; a mixed sequence of text and audio tokens is fed as input and the model decodes text or audio tokens.
    Audio tokens are converted back to raw audio with the latter AudioLM stages or SoundStorm (see Section~\ref{subsec:audiolmstages23}).}
    \label{fig:audio-palm-model}
\end{figure}

We use a decoder-only Transformer to model sequences consisting of text and audio tokens.
As far as the model is concerned, text and audio are just sequences of arbitrary integers, as the inputs are tokenized before feeding to the model, and any outputs are detokenized before being returned to a user of the model.
By representing speech with discrete tokens in a finite vocabulary, we can build a multimodal vocabulary which is the union of this audio vocabulary and a SentencePiece~\citep{sentencepiece} one used to represent text.
Thus, in principle there is almost no difference between our setting and the usual decoder-only setup for pure text, except that in our setting some of the tokens represent audio and some text, and we initialize our multimodal model using a pretrained text-only checkpoint.

The overall model is described in Figure~\ref{fig:audio-palm-model}. In the rest of this section we describe the main steps of the model: first, how text and audio inputs are tokenized; second, how we modify existing pretrained text decoders to also model audio; and third, how we convert the model output into raw audio. Since the first and third steps are identical to the process used by \cite{borsos2022audiolm} and \citep{borsos2023soundstorm}, we keep our explanation of these points high-level and refer the reader to those papers for further details.

Finally, we describe how we finetune AudioPaLM on a mixture of combined speech and text tasks including speech recognition and translation from or into either speech or text.

\subsection{Audio Embeddings and Tokenization\label{subsec:method-tokenization}}

We follow the process of \cite{gslm, borsos2022audiolm} to convert raw waveforms into tokens. This involves extracting embeddings from an existing speech representation model and subsequently discretizing those embeddings into a limited set of audio tokens.
\cite{borsos2022audiolm} extract embeddings from the w2v-BERT model \citep{w2vbert} and quantize them via k-means. In this work, we experiment with the following approaches to obtain a set of discrete audio tokens.

\begin{itemize}
    \item \textbf{w2v-BERT}: We follow the procedure described in~\cite{borsos2022audiolm} with two modifications.
    First, we use a w2v-BERT model that has been trained on multilingual data, as opposed to the English-only setting of \cite{borsos2022audiolm}.
    Second, we do not normalize the embeddings before performing the k-means clustering. While \cite{borsos2022audiolm} found that the normalization removed speaker-identity information without degrading performance, we found in the multilingual setting that normalization did indeed cause degradation.
    This method produces tokens at a rate of 25Hz and the token vocabulary is of size 1024. 
    \item \textbf{USM-v1}: We perform the same procedure with the more performant \emph{Universal Speech Model (USM)} encoder \citep{zhang2023usm} instead of the w2v-BERT encoder. 
    We use the largest 2B parameter variant of this multilingual speech encoder and extract embeddings from the middle layer.
    Similar to w2v-BERT, this method produces tokens at a rate of 25Hz and the token vocabulary is of size 1024.
    \item \textbf{\usmvtwo}: We additionally experiment with a quantizer that is trained with an auxiliary ASR loss.
    This version has been finetuned further to provide better multilingual performance. 
    As with USM-v1, this method accepts raw audio as input and returns a sequence of integers with length proportional to the length of the audio as output.
\end{itemize}

\subsection{Modifying text-only decoders to model both text and audio\label{subsec:method-checkpoint-surgery}}

In a Transformer decoder, the first layer of the model after input preprocessing is the \emph{token embeddings matrix} $\mathbf{E}$ which maps integer-valued tokens to dense embeddings; given a vocabulary of $t$ tokens and embeddings of size $m$, $\mathbf{E}$ is a $t \times m$ matrix whose $i$th row gives the embedding for the $i$th token.
Another embeddings matrix $\mathbf{E}'$ appears in the final softmax layer used to compute the logits over all tokens at each position; it is a $m \times t$ matrix which is multiplied with the $m$-dimensional output of the model to obtain a $t$ dimensional vector of logits, one for each of the tokens.
In the PaLM architecture, these matrices have shared variables, so that one is the transpose of the other, that is, $\mathbf{E}' = \mathbf{E}^\intercal$.

The rest of the decoder architecture is completely agnostic to the number of tokens modelled. 
Therefore we only need to make one small modification to turn a text-only model into one that models both text and audio: we expand the size of the embeddings matrix $\mathbf{E}$ to be of size $ (t + a) \times m $ where $a$ is the number of audio tokens (the size of $\mathbf{E}' = \mathbf{E}^\intercal$ changes accordingly). 

In order to make use of pretrained text models, we change the existing model checkpoints by adding $a$ new rows to the embeddings matrix $\mathbf{E}$. An implementation detail is that the first $t$ tokens (from zero to $t$) correspond to the SentencePiece text tokens while the next $a$ tokens (from $t$ to $t+a$) represent audio tokens.
While we can re-use the text embeddings of the pre-trained model, the new audio embeddings are freshly initialized and must be trained. We found it necessary to train all model parameters rather than keeping the previous weights fixed. We train using mixed speech and text tasks, as detailed in Section~\ref{sec:data-etc}. In Section~\ref{subsec:exp-from-scratch-vs-finetune} we show how adding audio tokens to a text-pretrained checkpoint in the above manner is highly beneficial for performance on the considered speech and text tasks (compared to re-training from scratch).
For further details about the PaLM architecture we refer the reader to Section 2 of \citep{chowdhery2022palm}.

\subsection{Decoding audio tokens to raw audio}
\label{subsec:audiolmstages23}

To synthesize an audio waveform from audio tokens, we experimented with two different methods: i) autoregressive decoding, following the setup of AudioLM~\citep{borsos2022audiolm} and ii) non-autoregressive decoding, using the recently proposed SoundStorm model~\citep{borsos2023soundstorm}. In both cases the audio tokens are first used to generate SoundStream tokens~\citep{zeghidour2021soundstream}, which are then converted to an audio waveform with a convolutional decoder.

The acoustic generation in AudioLM proceeds in two stages: ``Stage 2'' is a decoder-only Transformer model that takes the audio tokens produced by AudioPaLM and a voice conditioning as input, and generates SoundStream tokens that can be used to materialize the speech in the desired voice, but at a very low bitrate. ``Stage 3'' reconstructs higher levels of SoundStream's residual vector quantizer, which increases the bitrate and improves the audio quality. We use the same hyperparameters and the training process as in~\citep{kharitonov2023speartts}.

SoundStorm proposes an alternative non-autoregressive decoding scheme, which applies an iterative method that proceeds in parallel on all tokens. SoundStorm produces audio of the same quality as AudioLM, but with higher consistency in voice and acoustic conditions, while being two orders of magnitude faster.

In both cases we train on Multilingual LibriSpeech~\citep{mls} and the voice conditioning is supplied as a 3-second long voice sample, represented as both audio tokens and SoundStream tokens. 
By providing part of the original input speech as the voice conditioning, the model is able to preserve the original speaker's voice when translating their speech to a different language (see Section~\ref{sec:experiments}). Whenever the original audio is shorter than 3 seconds, it is repeated to reach the required duration. 

\subsection{Training tasks\label{subsec:method-tasks}}

\paragraph{Types of tasks}

We apply our method to the problems of speech recognition, speech synthesis and speech-to-speech translation.
All datasets used in this report are speech-text datasets which contain a subset of the following fields.
\begin{itemize}
    \item Audio: speech in the source language.
    \item Transcript: a transcript of the speech in Audio.
    \item Translated audio: the spoken translation of the speech in Audio.
    \item Translated transcript: the written translation of the speech in Audio.
\end{itemize}

The component tasks that we consider in this report are:
\begin{itemize}
    \item ASR (automatic speech recognition): transcribing the audio to obtain the transcript.
    \item AST (automatic speech translation): translating the audio to obtain the translated transcript.
    \item S2ST (speech-to-speech translation): translating the audio to obtain the translated audio.
    \item TTS (text-to-speech): reading out the transcription to obtain the audio.
    \item MT (text-to-text machine translation): translating the transcript to obtain the translated transcript.
\end{itemize}

A dataset including more than two of the fields may be used for multiple possible tasks.  As explored in the experiment of Section~\ref{subsec:exp-multiple-tasks}, we found that including multiple tasks (for example, both ASR and AST) from the same dataset resulted in improved performance.

\paragraph{Expressing tasks}
Following \cite{raffel2020exploring}, we signal to the model which task it should perform on a given input by prefixing the input with a tag specifying the task and the English name of the language of the input and, optionally, the language of the output if it is different. 

For example, to query the model to perform ASR on an utterance in French, the tokenized audio input would be preceded by the tag \texttt{[ASR French]}.
To perform TTS in English, the text would be preceded by \texttt{[TTS English]}.
To perform S2ST from English to French, the tokenized English audio would be preceded by \texttt{[S2ST English French]}.
The tag is tokenized using the normal text tokenizer of the model; we do not introduce special tokens to express the task or the languages involved. We found that changing task names to be more human-readable, such as using \texttt{transcribe the following French audio} instead of \texttt{[ASR French]}, does not change the performance of the model. Naming the language in the task -- compared to just using generic tags like \texttt{transcribe audio} or \texttt{[ASR]} -- is not ultimately required but is beneficial for low-resource languages.

\paragraph{Combined tasks}

We consider both \emph{direct tasks}, where the model is expected to directly map from input to output, and \emph{combined tasks}, where we instruct the model to also output intermediate steps for a complex task. This is similar in spirit to \emph{chain of thought prompting} \citep{wei2022chain}.

For example, for S2ST we could demand that the model directly maps from English audio tokens to French audio tokens. 
This would be expressed with the task tag \texttt{[S2ST English French]}.
Alternatively we can train the model to first output English text, followed by French text, and finally French audio tokens. 
We express this with the task tag \texttt{[ASR AST S2ST English French]}.
The model performs this task as a single autoregressive decoding, i.e. it is not performed with multiple separate calls to the model for each task. 
In particular this means that the model can attend to the input and all prior decoded content at each stage, as opposed to a separated pipeline approach of doing ASR, MT and then TTS.

We found combined tasks to improve performance, which we explore in the experiment of Section~\ref{subsec:exp-task-phrasing}.

\subsection{Training mixtures\label{subsec:method-mixtures}}

In this section we describe the data mixtures used to train our best models based on the datasets listed in Table~\ref{table:data}. Mixtures were implemented using the SeqIO library~\citep{roberts2022scaling}. More details on the datasets can be found in Section~\ref{sec:data-etc}.

There are two mixtures: one used to train the \emph{Audio PaLM 8B AST} and \emph{AudioPaLM-2 8B AST} models which output text and are trained on ASR and AST tasks; the other used to train the \emph{Audio PaLM 8B S2ST} model which outputs both text and speech and additionally includes TTS and S2ST tasks.

\begin{itemize}
    \item The \emph{AST mixture} is composed of:
    \begin{itemize}
        \item The ASR tasks from the CVSS, VoxPopuli ASR, CommonVoice 11, Conversational EsEn and Youtube ASR datasets. For the CVSS and Conversational EsEn datasets, we use ASR in both source and target languages.
        \item The AST tasks from CVSS, Conversational EsEn and VoxPopuli S2ST. We use VoxPopuli S2ST for AST by mapping from the translated audio to the transcript, since the translated transcript is not available. 
        \item The combined AST + ASR task for the CVSS and Conversational EsEn datasets.
        \item The MT task from the WMT/TED dataset. 

    \end{itemize}
    \item The \emph{S2ST mixture} is composed of the above, plus additionally:
    \begin{itemize}
        \item The TTS tasks from the CVSS and VoxPopuli ASR datasets. For CVSS we use only the source transcript and audio.
        \item The S2ST tasks from the Vox Populi S2ST, CVSS, WMT/TED and PaLM MT TTS datasets. Note that except for VoxPopuli S2ST, the speech targets of these datasets are all synthetically generated. For VoxPopuli S2ST we perform translation from both source to target, and target to source.
        \item The combined ASR + AST + S2ST tasks from the Conversational EsEn, CVSS and WMT/TED datasets.
    \end{itemize}
    
\end{itemize}

In general the components of the mixture are weighted according to the number of elements in each component while we downweighted larger datasets; Table~\ref{table:data} lists the amounts of audio that models trained on the above mixtures have seen during training. 

\subsection{Training setup}
In all experiments, we use the same finetuning setup as described in Section 6.1.2 of~\citep{chowdhery2022palm}.  
In particular, we finetune with the Adafactor optimizer with a constant learning rate of $5 \times 10^{−5}$ and dropout rate of 0.1, and we use loss masking on the inputs.

\section{Data and Metrics \label{sec:data-etc}}

\subsection{Datasets}

\begin{table}
  \caption{Datasets used for training AudioPaLM. The number of training hours corresponds to the number of hours seen by the AudioPaLM AST, AudioPaLM-2 AST and AudioPaLM S2ST models during training, as a result of datasets balancing and a finite number of training steps.}
  \label{table:data}
  \centering
  \setlength\tabcolsep{4pt}
  \noindent\makebox[1\textwidth]{
  \small
  \begin{tabular}{lcccccccl}
    \toprule
    \multirow{2}{*}{Name}     & \multirow{2}{*}{Audio} & \multirow{2}{*}{Transcript} & Translated & Translated & \multicolumn{3}{c}{\# Hours of training audio}& \multirow{2}{*}{\# Languages} \\
      &  &  & audio & transcript   & AudioPaLM AST & AudioPaLM-2 AST & AudioPaLM S2TS& \\
    \midrule
    CoVoST2 / CVSS & \checkmark  & \checkmark  & \checkmark  & \checkmark & 0.9k & 0.9k & 1.4k & 21 pairs X $\rightarrow$ En    \\ %
    VoxPopuli ASR & \checkmark  & \checkmark  & -  & -  & 1.6k & 1.6k  & 1.6k & 14    \\ %
    VoxPopuli S2ST & \checkmark  & \checkmark  & \checkmark  & -   & 6.7k & 3.9k & 5.4k & 15    \\ %
    CommonVoice 11 & \checkmark  & \checkmark  & -  & -  & 3.4k & 2k  & 0.9K & 98    \\
    Conversational EsEn & \checkmark  & \checkmark  & \checkmark  & \checkmark  & 2k & 2k & 2k & 2   \\
    YouTube ASR & \checkmark  & \checkmark  & -  & -  & 13k & 7.6k & - & 56    \\
    WMT/TED TTS & \checkmark  & \checkmark  & \checkmark  & \checkmark  & - & - & 26.2k & 21 pairs X $\rightarrow$ En    \\ %
    PaLM MT TTS & \checkmark  & \checkmark  & \checkmark  & \checkmark  & - & - & 10.9k &  113 pairs X $\rightarrow$ En     \\ %
    \bottomrule
  \end{tabular}
  }
\end{table}

Table~\ref{table:data} lists the datasets used in AudioPaLM training.
\begin{itemize}
  
\item CoVoST2~\citep{wang2022covost2} is a speech-to-text dataset mapping speech in 21 languages to English text.
\item CVSS~\citep{jia2022cvss} augmented CoVoST2 to synthesize speech for the target text in two flavors: CVSS-C uses a canonical speakers voice, while CVSS-T transfers voice properties from the source voice. Unless stated otherwise, we use the CVSS-C flavor in speech-to-speech translation experiments.
\item VoxPopuli~\citep{wang2021voxpopuli} contains speeches from the European Parliament together with their transcripts -- which can be used for speech recognition (ASR) tasks -- as well as spoken translations from parliamentary interpreters -- which can be used for speech translation (S2ST) tasks.
\item Common Voice~\citep{commonvoice} consists of text paired with recordings where people were asked to read the text aloud.
\item The conversational dataset described in \citep{jia2019leveraging} was obtained by crowd-sourcing humans to read a subset of the Spanish side of a proprietary conversational Spanish-English MT dataset.
\item YouTube ASR is an unlabeled multilingual dataset of YouTube-based audio which was transcribed automatically by using the USM-2B ASR model~\citep{zhang2023usm}. The dataset helps to improve models for YouTube to perform better on captioning and translation.

\item WMT/TED TTS is based on WMT  \citep{barrault2020findings,barrault2019findings,bojar2018findings,bojar2017findings,bojar2015findings,bojar2013findings} and TED \citep{qi2018whenwhy} text-to-text translation datasets as described in~\citep{bapna2022mslam}. Following~\citet{jia2022leveraging} the dataset is augmented by running all the source and target text through a TTS engine to generate synthetic paired audio.

\item PaLM MT TTS provides additional training data for S2ST: We use PaLM-2 to translate the transcripts of the YouTube, Common Voice, and Babel~\citep{gales2017babel} datasets to English, and use a prior AudioPaLM 8B S2ST model (trained without this dataset) to synthesize the speech. The method is similar in spirit to~ \citep{jia2019leveraging} which combines MT and TTS to generate additional paired data.

\end{itemize}

We train models on mixtures based on these datasets as described in Section~\ref{subsec:method-mixtures} on ASR, AST, and S2ST tasks from the above datasets.
In Section~\ref{subsec:exp-scale-train-data} we explore how adding more data improves the performance of our method.

Note that our method makes use of the text-pretrained PaLM checkpoints and audio tokenizers. So while the models are trained on the datasets listed in Table~\ref{table:data}, they can also benefit from PaLM's text training data~\citep{palmv2} via the pre-trained PaLM checkpoint, and from the data used to train the audio tokenizers.

\subsection{Evaluation Metrics}

We evaluate our method on the following benchmarks:

\begin{itemize}
    \item CoVoST2 AST: We use BLEU scores, with the SacreBLEU corpusBLEU implementation \cite{papineni2002bleu, post-2018-call}. We do not perform any normalization to the text before computing BLEU.
    \item FLEURS AST: The FLEURS~\citep{conneau2023fleurs} dataset contains speech utterances and their corresponding transcripts in 102 languages and is used for evaluation, only. We use BLEU scores, as described for CoVoST2 AST.
    \item VoxPopuli ASR: We use the JiWER implementation of word error rate (WER). We normalise the text by ignoring capitalisation and punctuation before computing the WER.
    \item CoVoST2 ASR: Comparable to VoxPopuli ASR, but for Japanese and Chinese, the character error rate (CER) is reported instead of WER. We report this metric for experiments trained on CoVoST2, only.
    \item CVSS S2ST: 
    Following Translatron 2 \citep{jia2022translatotron}, we feed the audio output of our model into an ASR model and use BLEU to compare the ASR output with the ground truth target text. 
    We use the same ASR model as \cite{jia2022translatotron} and so the metrics presented here are directly comparable.
\end{itemize}

All evaluations are performed on the test splits of the corresponding datasets.

\section{Experiments\label{sec:experiments}}

We start with our top-level results presenting significant improvements over prior results on automatic speech-to-text translation (AST) and direct speech-to-speech translation (S2ST), as well as competitive results on automatic speech recognition (ASR).
Ablations of individual factors are provided in Section~\ref{subsec:experiments-ablations}.

\subsection{Speech translation and speech recognition results\label{subsec:toplevel_results}}

Table~\ref{table:top-level-results} displays results on ASR, AST and S2ST benchmarks for our method and existing baselines.
Our models come in two variants; the first variant (referred to as AST) is trained on AST tasks \emph{without} S2ST and TTS data; the second variant (referred to as S2ST) is trained \emph{with} S2ST and TTS data and is therefore able to produce speech as well. To generate the audio for the S2ST results we used SoundStorm~\citep{borsos2023soundstorm}.
For details on the training mixtures see Section~\ref{subsec:method-mixtures}. 

As an initial checkpoint, we use a PaLM-2 8B checkpoint~\citep{palmv2} to which we add the capability to process audio tokens as input and output as described in  Section~\ref{subsec:method-checkpoint-surgery}. The additional audio token embeddings are initialized to $0$.
As in the original PaLM and PaLM 2 models, the input and output embeddings are shared.

\begin{table}
  \caption{Top level results of this paper.}
  \label{table:top-level-results}
  \begin{adjustwidth}{-.5in}{-.5in}  
  \centering
  \begin{tabular}{
  l
  d{3.1}
  d{3.1}
  d{3.1}
  }
    \toprule

    \multirow{2}{*}{Model} 
       & \multicolumn{1}{c}{CoVoST2 AST} & \multicolumn{1}{c}{CVSS S2ST} & \multicolumn{1}{c}{VoxPopuli ASR} \\
       & \multicolumn{1}{c}{BLEU$\uparrow$} & \multicolumn{1}{c}{ASR-BLEU$\uparrow$} & \multicolumn{1}{c}{WER$\downarrow$} \\
    \midrule
    \midrule
    Whisper Large-v2 1.5B \citep{radford2022whisper} & $29.1$  & - & $13.6$     \\
    mSLAM-CTC 2B \citep{bapna2022mslam} & $25.2$  & - & $9.1$   \\
    MAESTRO 600M \citep{chen2022maestro} & $25.2$  & - & \textbf{8.1}     \\
    USM-M \citep{zhang2023usm} & $30.7$ & - & - \\
    Translatotron 2 + pretraining  & \multirow{2}{*}{$-$}  & \multirow{2}{*}{25.6} & \multirow{2}{*}{$-$}    \\
    \enspace  + TTS aug \citep{jia2022leveraging} & & & \\
    \midrule
    AudioPaLM 8B AST (ours) & $35.4$  & - & $11.1$       \\
    AudioPaLM 8B S2ST (ours) & $36.2$  & \textbf{32.5} & $16.0$        \\
    \midrule
    AudioPaLM-2 8B AST (ours) & \textbf{37.8}   & - & $9.8$       \\
    \midrule
    AudioPaLM-2 8B cascaded ASR + transl. (ours)  & $39.0$  & -   & -     \\
    \bottomrule
  \end{tabular}
  \end{adjustwidth}
\end{table}

Our method exceeds the baselines on AST and S2ST and is competitive on ASR.
Our method also comes close in AST performance to a cascaded approach in which we use our best AudioPaLM-2 ASR model followed by translation with another AudioPaLM-2 model finetuned only for text-to-text translation on CoVoST2.

\subsection{Zero-shot behaviour\label{subsec:zeroshot_results}}
\paragraph{Setup.} We evaluate the zero-shot capabilities of our AST models on the FLEURS multilingual dataset~\citep{conneau2023fleurs}. The dataset contains speech utterances and their corresponding transcripts in 102 languages. Note that none of our models were trained on FLEURS, so we use the dataset for evaluation only. In this context, we focus on the language pairs X $\rightarrow$ English and extract two subsets of languages:
\begin{itemize}
    \item 29 \emph{AST-observed languages}: languages for which speech-to-text translation (AST) data (X $\rightarrow$ En) was seen during training (as these language pairs were present in the VoxPopuli S2ST, CoVoST2 or/and Conversational EsEn datasets). These languages are indicated with a \S~in Table~\ref{table:fleurs-bleu-per-language}.
    \item 26 \emph{ASR-observed languages}: languages for which no speech-to-text translation data was seen when training our AST models, but for which at least 1 hour of transcription (ASR) data was present. We removed 3 languages (Cantonese, Kurdish and Ganda) for which we did not have a BLEU score for the baseline. These languages are indicated with a {\textdagger} in Table~\ref{table:fleurs-bleu-per-language}.
\end{itemize}

\paragraph{Results.} In Table~\ref{table:zero-shot-s2t}, we present the results obtained with the two proposed AST models AudioPaLM and AudioPaLM-2, as well as the baseline model ``Whisper Large-v2 1.5B''. We also present the number of AST and ASR speech training hours for these three models. For the proposed models, the reported number of hours do not take into account the amount of speech used to train the tokenizers. 

\paragraph{Discussion.} We observe that the proposed AudioPaLM-2 model  significantly outperforms the Whisper model on AST-observed languages.
Although Whisper is used as a reference for the \emph{only ASR observed} setting, its results are not zero-shot as Whisper has been trained on 40.6K hours of speech-to-text translation (AST) data for these languages.
For the AudioPaLM models, this setting is zero-shot as it did not see any AST data for these languages.
Despite this disadvantage, AudioPaLM-2 also outperforms the Whisper model on ASR-observed languages. 
For a detailed performance comparison for each language, see Appendix~\ref{appendix:ast-zero-shot}.

There is a large improvement obtained by using the AudioPaLM-2 instead of the AudioPaLM model: 28\% increase for AST-observed languages and 107\% increase for ASR-observed languages. These numbers show that the superior text translation capabilities of AudioPaLM-2 immediately transfer to the audio domain, despite the fact that the model has not seen any speech-to-text data for these language pairs during training in the case of ASR-observed languages.

\begin{table}
  \caption{Zero-shot AST performance on the FLEURS dataset. We split the results into two groups: \emph{AST observed} are languages for which X$\rightarrow$ En speech-to-text data was present in the AudioPaLM training data. \emph{Only ASR observed} are languages for which only ASR data was present in the training data (so the translation is done by the AudioPaLM model in a zero-shot manner). This shows that AudioPaLM inherits its translation capabilities from the base model, and is consistent with the improved translation capabilities of PaLM-2 compared to PaLM. The hours of training data for AudioPaLM do not include audio data seen in self-supervised training of audio tokenization models.
  \vspace{1mm} \newline
  $^*$Whisper has seen AST data for all languages considered and is included here just for reference. 
  }
  
  \label{table:zero-shot-s2t}
  \centering
  \noindent\makebox[1\textwidth]{
  \begin{tabular}{
    lcccc
  }
    \toprule
    \multirow{2}{*}{Model} 
       & \multicolumn{2}{c}{\textbf{AST observed languages}} &  \multicolumn{2}{c}{\textbf{Only ASR observed languages}}  \\
       & \multicolumn{1}{c}{BLEU$\uparrow$} & AST / ASR hours &  \multicolumn{1}{c}{BLEU$\uparrow$} & AST / ASR hours \\
    
    \midrule
    Whisper Large-v2 1.5B \citep{radford2022whisper} & $23.3$ &  74.0k / 104.4k &  $19.6^*$ & 40.6k / 11.3k \\
    \midrule
    AudioPaLM 8B AST & $22.4$ & 6.6k / 11.4k & $10.0$\phantom{)}  &  0 / 5.3k \\
    AudioPaLM-2 8B AST & \textbf{28.6} & 4.8k / 8.2k & \textbf{20.7}\phantom{)} & \textbf{0 / 3.1k}      \\
    \bottomrule
  \end{tabular}}
\end{table}

\subsection{Quality of generated speech}
In addition to measuring the translation quality of the speech content as reported in Table~\ref{table:top-level-results}, we are also interested in evaluating whether the speech generated by \ours is (a) of high quality, and (b) truthfully preserves the voice of the speaker when translating to a different language. To this end, we use a combination of objective metrics and subjective evaluation studies that use the test split of the CVSS-T dataset~\citep{jia2022cvss}. The subjective experiments were conducted on an earlier version of AudioPaLM using the acoustic generation method described in AudioLM~\citep{borsos2022audiolm}. 

\paragraph{Baselines.} As a first baseline, we use the ground-truth translated utterances which are provided as a part of the CVSS-T dataset. These utterances were obtained by synthesizing the ground-truth translated text with a high-quality TTS system which was modified to enable voice transfer~\citep{jia2021png,jia2022cvss}. As a result, the ground-truth utterances mimic the voice in the source utterance. 

As a second baseline, we use Translatotron 2. Note that we could not use the  
``\ttron + pretraining + TTS aug'' model (mentioned in Table
~\ref{table:top-level-results}) in the comparison because it was not trained to preserve voices and instead generates speech in a single canonical voice. Instead, we use the \ttron system presented by~\citet{jia2022cvss} which is capable of transferring the voice from the source utterance (albeit it achieves a lower BLEU score on CVSS). This version of the \ttron model was trained on the CVSS-T dataset and implements S2ST from 21 languages to English.

\paragraph{Objective metrics.}
\looseness=-1
As the first objective metric, we use a no-reference MOS estimator akin to~\cite{dnsmos} which, given an audio sample, provides an estimate of the perceived audio quality on a scale from 1 to 5. To measure cross-lingual voice transfer quality, we rely on an off-the-shelf speaker verification model~\citep{wavlm} as used by~\cite{vallex} and \cite{kharitonov2023speartts}, and compute the cosine similarity between the embeddings of the source (encoded/decoded with SoundStream) and the translated speech. Besides voice preservation, we also measure how well the acoustic properties (recording conditions, background noise) are transferred from the source audio to the target. We do so by computing the cosine similarity between embeddings extracted from a model trained to identify segments that belong to the same recording~\citep{borsos2023soundstorm}. %

\paragraph{Subjective evaluation.} We run two separate studies, one for evaluating the quality of the generated speech, and another for assessing the voice similarity. We use the same set of samples for both studies. Since utterances in CVSS-T are sourced from volunteer-generated data of variable quality, we noticed that some of the utterances contain loud overlapping speech (e.g., a TV show or a song playing in the background) or extremely strong noise (e.g., clothes rubbing against the microphone). 
Such aberrations complicate the work of raters, thus we decided to pre-filter by only selecting inputs with an estimated MOS of at least $3.0$. Finally, we sampled 10 examples per language, giving us $21 \times 10 = 210$ source utterances to translate. All utterances were peak normalised and resampled to 16kHz, if needed.

Before starting, the raters were provided with a small set of illustrative examples with ground-truth grades. They also completed a small pilot study as a training. The utterances (pairs of source-target utterances, in the case of the voice similarity evaluation) were presented one-by-one. The ratings are provided on a 5-grade scale from 1 (poor quality or completely different voices) to 5 (excellent quality, identical voices). In the voice similarity study, the raters are explicitly asked to ignore differences in the recording conditions and language, and solely focus on the voice. Each of the $630$ output examples ($10$ inputs from each of $21$ languages were generated with each of the $3$ different systems) was rated $10$ times which results in 6300 ratings per study. %
Aggregating those ratings per system, we obtain mean opinion score (MOS) and similarity mean opinion score (SMOS).

\paragraph{Results.} We report the results of the objective and subjective evaluations in Table~\ref{table:speech-eval}. From these results we observe that \ours significantly outperforms the baseline \ttron system both in audio quality and in voice similarity, in objective and subjective measurements. Moreover, \ours has higher quality and better voice similarity than the ground-truth synthesized recordings in CVSS-T, with a relatively large gap in most of the metrics. 
Following~\cite{jia2022cvss}, we also compared the systems across high and low-resource groups (French, German, Spanish and Catalan vs.\ the rest) and found no significant variation of the metrics across these groups.

\begin{table}
  \caption{Audio quality and voice similarity results. Subjective and objective audio quality results are reported on the 1...5 MOS scale. Objective voice similarity and acoustic consistency are measured in terms of cosine similarity. Subjective voice similarity scores span 1...5. Both objective and subjective metrics are computed on the same set of examples. Higher is better across all metrics.}
  \label{table:speech-eval}
  \centering
  \begin{tabular}{cccccc}
    \toprule
      & \multicolumn{2}{c}{\textbf{Audio quality}}  &  \multicolumn{2}{c}{\textbf{Voice similarity}} & \multicolumn{1}{c}{\textbf{Acoustic consistency}}\\
     & \multicolumn{1}{c}{Objective} & \multicolumn{1}{c}{Subjective}  & \multicolumn{1}{c}{Objective} & \multicolumn{1}{c}{Subjective} & \multicolumn{1}{c}{Objective} \\
     \midrule
    CVSS-T & 3.41 & 3.88 & 0.24 & 3.70 & 0.54\\
    
    \midrule
    \ttron & 3.36 & 3.96 & 0.18 & 3.51 & 0.44 \\
    \ours & \bf{3.65} & \bf{4.44} & \bf{0.40} & \bf{4.00} & \bf{0.81} \\

    \bottomrule
  \end{tabular}
\end{table}

\subsection{Impact of model and data choices}\label{subsec:experiments-ablations}

In this section we walk the reader through experiments that guided us towards our final training recipe from initial early experimentation.
These show the impact of individual factors and build on top of one another until reaching the final setup described and analysed in the previous sections.

\subsubsection{Training on multiple tasks\label{subsec:exp-multiple-tasks}}

To achieve the results in Section~\ref{subsec:toplevel_results}, we trained on multiple tasks based on the same underlying data to improve performance. For example, the CoVoST2 data can be used for both ASR and AST tasks, and we observed that adding ASR tasks in training results in improved performance on AST benchmarks, compared to training with the AST tasks alone. In this section we investigate the effect of this choice on model performance.

\paragraph{Setup.}
We train two models on the CoVoST2 dataset. 
All conditions are identical except that in one experiment, we use only the AST data; in the other we train with both AST and ASR tasks. 
The base models are the PaLM 8B checkpoint and we use the USM-v1 tokenizer.
We evaluate on the CoVoST2 AST benchmark.

\paragraph{Results.}
See Table~\ref{table:multiple-tasks}.
We observe that adding ASR tasks into the dataset increases BLEU by 2.5 from 16.0 to 18.5 on the CoVoST2 AST benchmark.

\begin{table}
  \caption{Results from experiment \ref{subsec:exp-multiple-tasks} showing the impact of training with ASR data in addition to AST data. Adding ASR tasks to the training mix helps to improve performance on AST.}
  \label{table:multiple-tasks}
  \centering
  \begin{tabular}{lcc}
    \toprule
    \multirow{2}{*}{Tasks} & CoVoST2 AST  \\
    & BLEU$\uparrow$ \\
    \midrule
    AST only  & 16.0  \\
    AST \& ASR & 18.5  \\
    \bottomrule
  \end{tabular}
\end{table}

\paragraph{Discussion.}
Although ASR is not part of the evaluation task, adding ASR data helped improve performance. 
Our hypothesis is that ASR tasks help the model to better connect its understanding of the new audio input to its previous understanding of text.
In subsequent experiments we include both ASR and AST tasks when using the CoVoST2 training data.

\subsubsection{Training from scratch vs. finetuning\label{subsec:exp-from-scratch-vs-finetune}}

The results in Section~\ref{subsec:toplevel_results} are based on finetuning a text-pretrained PaLM checkpoint. Here we investigate the effect of using such a model compared to starting training from scratch on the same architecture. 

\paragraph{Setup.} 
 
In the \emph{1B from-scratch} and \emph{8B from-scratch} experiments we start with randomly initialised weights. 
In the \emph{8B finetune} experiment we start from the PaLM 8B checkpoint, which has been modified by adding extra rows to the token embedding matrix for the audio tokens, which are randomly initialised.

All three models are trained on CoVoST2 ASR and AST tasks.

\paragraph{Results.} 
See Table~\ref{table:from-scratch-vs-finetune}.
We observe that finetuning the PaLM 8B checkpoint achieves substantially higher performance than training from scratch on CoVoST2 tasks for both ASR and AST. The \emph{1B-from-scratch} experiment was added to determine whether a smaller model architecture would work better than the \emph{8B} model when trained from scratch on CoVoST2; it does not.

\begin{table}
  \caption{Results from Experiment \ref{subsec:exp-from-scratch-vs-finetune} showing that training from a pretrained checkpoint has a substantial positive effect on performance compared to training from scratch.}
  \label{table:from-scratch-vs-finetune}
  \centering
  \begin{tabular}{
     l
     d{3.1}
     d{3.1}
  }

    \toprule
    \multirow{2}{*}{Initial checkpoint} 
    & \multicolumn{1}{c}{CoVoST2 AST} & \multicolumn{1}{c}{CoVoST2 ASR}  \\
    & \multicolumn{1}{c}{BLEU$\uparrow$} & \multicolumn{1}{c}{WER$\downarrow$} \\
    \midrule
    PaLM 1B from scratch & $6.5$ & $66.0$ \\
    PaLM 8B from scratch & $6.9$  & $63.3$       \\
    PaLM 8B finetuned & $18.4$  & $40.2$       \\
    \bottomrule
  \end{tabular}
\end{table}

\paragraph{Discussion.}
Finetuning a pretrained checkpoint substantially improves results.
This is in some sense not surprising as the base model is very capable to begin with; nonetheless it is interesting that with finetuning the model is able to adapt to completely new input stimulus, since the audio tokens are totally new embeddings that the model must learn to understand.
Furthermore the audio tokens are very different from text: despite the low sampling rate, there is presumably still some redundancy in the data and the rate of samples is still much higher than text tokens --- we estimate from the data that at 25Hz, one text token corresponds to approximately 6-8 audio tokens.

\subsubsection{Different tokenization schemes
\label{subsec:exp-tokens}}

To obtain the results in Section~\ref{subsec:toplevel_results}, we tokenized audio based on USM-v2. Here we investigate the impact of the choice of tokenization scheme on the final results. 

\paragraph{Setup.}
We train three models with all conditions identical except for the tokenization scheme applied to the audio. All models are trained using the PaLM 8B checkpoint.
In each case we use the CVSS datasets with ASR and AST tasks with the source audio preprocessed using different tokenizers.
The three tokenizers used are the w2v-BERT, USM-v1 and \usmvtwo tokenizers which were discussed Section~\ref{subsec:method-tokenization}.

\paragraph{Results.}

See Table~\ref{table:exp-tokens}. 
We observe that the choice of tokenization scheme has a large impact on the performance of the model. 
The fact that the USM encoder is more powerful than w2v-BERT indeed translates to an improvement in performance in our setting.
The \usmvtwo tokens perform even better, yielding substantially improved results.

\begin{table}
  \caption{Results from Experiment \ref{subsec:exp-tokens} showing the impact of training with different types of tokens. Performance is affected significantly by the choice of tokens.}
  \label{table:exp-tokens}
  \centering
  \begin{tabular}{
  l
  d{3.1}
  d{3.1}
  }
    \toprule
    \multirow{2}{*}{Tokens} 
    & \multicolumn{1}{c}{CoVoST2 AST} & \multicolumn{1}{c}{CoVoST2 ASR}  \\
    & \multicolumn{1}{c}{BLEU$\uparrow$} & \multicolumn{1}{c}{WER$\downarrow$} \\
    \midrule
    W2V-BERT    & $15.2$  & $50.1$       \\
    USM-v1         & $18.5$  & $40.2$       \\
    \usmvtwo   & $26.9$  & $22.3$       \\
    \bottomrule
  \end{tabular}
\end{table}

\paragraph{Discussion.}
The choice of tokenization scheme has a substantial effect on performance. 
This is not surprising; the model only is exposed to the information captured by the tokenizer, and this may be in a form which is easy or difficult for the model to process.
Future work should consider tokenization of audio more carefully because this is still relatively immature as a research area.

\subsubsection{Training with combined tasks
\label{subsec:exp-task-phrasing}}

To obtain the results in Section~\ref{subsec:toplevel_results}, we required the model to compute intermediate steps for complex tasks by combining multiple tasks into one, as described in Section~\ref{subsec:method-tasks}. In the following we investigate the impact of this choice.

\paragraph{Setup.}
We train pairs of models on the CoVoST2 AST dataset. 
Within a pair, the only change is that for one model we train with ASR and AST tasks, while for the other we also include the combined task consisting of first doing ASR and then outputting the AST result.
For the latter model, at evaluation time we report the result of doing the combined task from which we use only the final output.
We repeat this setup twice: once with the USM-v1 tokens, and once with the \usmvtwo tokens.

\paragraph{Results.} 
See Table~\ref{table:task-phrasing}. 
This shows that expressing the AST task as a combination of simpler tasks results in improved performance on the AST task.
At the same time, we see a small reduction in performance on the ASR task.

\begin{table}
  \caption{Results from Experiment \ref{subsec:exp-task-phrasing} showing that defining complex tasks as combinations of simpler tasks results in an improvement in performance on the AST task and a small reduction on the ASR task.}
  \label{table:task-phrasing}
  \centering
  \begin{tabular}{
    l
    l
    d{3.1}
    d{3.1}
    }
    \toprule
    \multirow{2}{*}{Tokens} &  \multirow{2}{*}{Tasks} 
    & \multicolumn{1}{c}{CoVoST2 AST} & \multicolumn{1}{c}{CoVoST2 ASR}  \\
    & & \multicolumn{1}{c}{BLEU$\uparrow$} & \multicolumn{1}{c}{WER$\downarrow$} \\
    \midrule
    \multirow{2}{*}{USM-v1}    & Direct     & $18.5$  & $40.2$   \\
                            & Combined    & $22.1$  & $41.6$   \\
    \midrule
    \multirow{2}{*}{\usmvtwo}    & Direct     & $26.9$  & $22.3$   \\
                                  & Combined    & $30.5$  & $25.3$     \\
    \bottomrule
  \end{tabular}
\end{table}

\paragraph{Discussion.}
Our results are consistent with prior works which have observed that allowing the model to break down a complex task into easier pieces results in improved performance, relative to making the model directly output the answer~\citep{wei2022chain}.

At the same time, we observe a reduction in performance on the ASR task.
We hypothesize that this may be a consequence of our checkpoint selection criterion, which was to select the checkpoint with the best AST metric on the validation split.
It may also be a consequence of the large change in the data mixture resulting from this change.

We note that it may appear that combined tasks reduce the problem to a pipeline approach of separate ASR and translation systems.
However this is not the case, as the model can refer to all previous tokens at each step and is a single unified model. 
For example, when decoding the translated text, it is possible to refer to the input audio and any information contained in them.
This is particularly important for the S2ST setting (see Experiment~\ref{subsec:exp-add-s2st}) where prosodic information may be present in the input audio, which can be attended to while decoding output audio.

\subsubsection{Training with additional speech-to-speech tasks
\label{subsec:exp-add-s2st}}

In the following, we investigate the impact of adding speech-to-speech translation (S2ST) tasks to the trained tasks.

\paragraph{Setup.}
We train two models using the CoVoST2 dataset. One model is only trained on the AST, ASR and combined AST tasks.
The other model is additionally trained on S2ST as a direct and combined task.
Thus the difference between these two models is that in the S2ST the model additionally sees tasks in which it must output audio tokens, whereas for the other tasks (and all previous experiments) the model only outputs text tokens.

\paragraph{Results.}
See Table~\ref{table:exp-add-s2st}. 
We observe that adding the S2ST task results in the new capability of being able to perform S2ST, but that this comes at the cost of a modest decrease in performance to both the AST BLEU score and ASR WER score when evaluating on the CoVoST2 test split.

\begin{table}
  \caption{Results from Experiment \ref{subsec:exp-add-s2st} showing that training with S2ST data brings new capabilities but results in a degradation of performance on AST and ASR tasks.}
  \label{table:exp-add-s2st}
  \centering
  \begin{tabular}{
    l
    d{3.1}
    d{3.1}
    d{3.1}
   }
    \toprule
    \multirow{2}{*}{Tasks} 
    & \multicolumn{1}{c}{CoVoST2 AST} & \multicolumn{1}{c}{CVSS S2ST} & \multicolumn{1}{c}{CoVoST2 ASR} \\
    & \multicolumn{1}{c}{BLEU$\uparrow$} & \multicolumn{1}{c}{ASR-BLEU$\uparrow$} & \multicolumn{1}{c}{WER$\downarrow$} \\

    \midrule
    AST, ASR         & $30.5$  & - & $25.3$ \\
    AST, ASR \& S2ST & $27.8$  & $24.2$ & $27.1$ \\
    \bottomrule
  \end{tabular}
\end{table}

\paragraph{Discussion.}
Since we use loss masking on the inputs for each training example, performing S2ST is fundamentally different from ASR or AST since the model must learn to emit audio tokens.
For ASR and AST, the model takes audio tokens as input, but the loss masking means that it doesn't need to learn to model these sequences of audio tokens.
It is thus perhaps not surprising that this results in a decrease in performance on the text-output tasks, since model capacity must be devoted to audio modelling.

\subsubsection{Scaling the training data
\label{subsec:exp-scale-train-data}}

In this section we investigate the impact of increasing the amount of training data.

\paragraph{Setup.}
We run this analysis on two types of models, both trained from a PaLM 8B checkpoint and with \usmvtwo tokens. The models ``AudioPaLM 8B AST'' are trained without the S2ST tasks, the models ``AudioPaLM 8B S2ST'' are trained with the S2ST tasks. 

We train these two types of models with an increasing amount of data: %
\begin{itemize}
    \item The CoVoST2 dataset only. For the S2ST model, we use the modified S2ST version of this dataset: CVSS.
    \item All the public datasets described in Table~\ref{table:data}, namely CoVoST2/CVSS, VoxPopuli AST, VoxPopuli S2ST, CommonVoice 11 and Conversational EsEn.
    \item All the public datasets, as well as the YouTube ASR dataset.
    \item All the public datasets, as well as the YouTube ASR dataset and the WMT/TED text-to-text translation dataset. For the S2ST models, we follow \cite{jia2022leveraging} and synthesise a paired S2ST dataset from this by using TTS on the examples in this dataset.
    \item As above, but using the synthetic PaLM-based MT TTS dataset S2ST mixture instead of the YouTube ASR dataset. For this dataset we used PaLM-2 to translate the transcripts of the YouTube, Common Voice, and Babel datasets to English text, and then synthesized the English speech to create a speech-to-speech dataset.
\end{itemize}

\paragraph{Results.}
See Table~\ref{table:scale-train-data}.
We observe that training with increasing amounts of data yields a substantial improvement. 
In particular, consistent with Experiment~\ref{subsec:exp-multiple-tasks} we see that adding additional ASR data helps on AST tasks.
Consistent with Experiment~\ref{subsec:exp-add-s2st} we observe that for each fixed dataset mixture for which we compare the AST and S2ST mixtures, including the S2ST tasks brings new S2ST capabilities at the cost of a modest reduction in performance on AST. All of the S2ST results in Table~\ref{table:scale-train-data} use AudioLM stage 2 and 3 models~\citep{borsos2022audiolm} to reconstruct the audio samples from audio tokens as discussed in Section~\ref{subsec:audiolmstages23}. 

\begin{table}
  \caption{Results for Experiment~\ref{subsec:exp-scale-train-data} showing that scaling the amount of training data improves performance. 
  Observe also that within each pair, adding S2ST tasks brings new capabilities, but at the expense of slight decrease in performance on AST and ASR tasks. ``Public speech datasets'' corresponds to CoVoST2/CVSS, Vox Populi, CommonVoice 11 and Conversational EsEn.}
  \label{table:scale-train-data}
  \begin{adjustwidth}{-.5in}{-.5in}
  \centering
  \begin{tabular}{
    l
    l
    d{3.1}
    d{3.1}
    d{3.1}
    d{3.1}
    }
    \toprule
    \multirow{2}{*}{Datasets}  & \multirow{2}{*}{Tasks}     
      & \multicolumn{1}{c}{VoxP. ASR}      &  \multicolumn{1}{c}{CoVoST2 ASR}  &  \multicolumn{1}{c}{CoVoST2 AST} & \multicolumn{1}{c}{CVSS S2ST}  \\
    & & \multicolumn{1}{c}{WER$\downarrow$} & \multicolumn{1}{c}{WER$\downarrow$} & \multicolumn{1}{c}{BLEU$\uparrow$} & \multicolumn{1}{c}{ASR-BLEU$\uparrow$} \\
    \midrule
    \multirow{2}{*}{CoVoST2 / CVSS} 
    &  AST & $168.7$ & $25.3$ & $30.5$  & -      \\
    & S2ST & $166.3$ & $27.1$ & $27.8$  & $24.2$ \\
    \midrule
    \multirow{2}{*}{Public speech datasets} 
     &  AST &  $9.0$  & $15.5$ & $33.1$ & -     \\
     & S2ST & $14.5$  & $19.4$ & $31.9$ & $27.0$    \\
    \midrule
    Public speech datasets &  AST &  $9.6$ & $13.8$ & $34.8$   & -      \\
     \enspace  + YT        & S2ST & $14.5$ & $16.5$ & $32.3$   & $27.6$ \\
    \midrule
    Public speech datasets    &  AST & $11.1$ & $15.1$ & $35.4$   & -   \\
     \enspace  + YT + WMT/TED & S2ST & $15.4$ & $16.6$ & $33.8$   & $29.5$     \\
    \midrule
    Public speech datasets    &  \multirow{2}{*}{S2ST} & \multirow{2}{*}{16.0} & \multirow{2}{*}{15.0} & \multirow{2}{*}{36.2}   & \multirow{2}{*}{31.2}   \\
     \enspace   + PaLM MT TTS + WMT/TED & & & & &      \\
    \bottomrule
  \end{tabular}
  \end{adjustwidth}
\end{table}

\paragraph{Discussion.}
It is unsurprising that scaling the amount of training data results in an improvement in performance.
We observe that adding more data in some cases leads to a small reduction in performance on the ASR tasks, though always an improvement on the AST tasks. Similar to Experiment~\ref{subsec:exp-task-phrasing}, this may be a consequence of our checkpoint selection criterion, which is based on AST performance on the CVSS validation set.

\subsubsection{Decoding with AudioLM vs SoundStorm
\label{subsec:exp-audio-lm-vs-soundstorm}}

In this section we investigate the impact on S2ST metrics of decoding using AudioLM stage 2 and 3 models vs SoundStorm.

\paragraph{Setup.}
We take the best AudioPaLM model from Section~\ref{subsec:exp-scale-train-data} trained with the mixture consisting of public, PaLM MT TTS and WMT/TED datasets.
The previous experiment used AudioLM stage 2 and 3 models to decode the audio tokens output by AudioPaLM to wave audio. We rerun this using a SoundStorm model instead, and measure the impact on the CVSS S2ST task.

\begin{table}
  \caption{Results for Experiment~\ref{subsec:exp-audio-lm-vs-soundstorm} showing the impact on S2ST metrics of decoding from audio tokens to wave audio using AudioLM stage 2 and 3 models compared to SoundStorm.}
  \label{table:audio-lm-vs-soundstorm}
  \centering
  \begin{tabular}{
    l
    d{3.1}
   }
    \toprule
    \multirow{2}{*}{Decoder} & 
    \multicolumn{1}{c}{CVSS S2ST}  \\
  & 
  \multicolumn{1}{c}{ASR-BLEU$\uparrow$} \\
    \midrule
    AudioLM  & 31.2\\
    SoundStorm  & 32.5\\
    \bottomrule
  \end{tabular}
\end{table}

\paragraph{Results.}
See Table~\ref{table:audio-lm-vs-soundstorm}.
We observe a 1.3 BLEU point increase when using SoundStorm compared to AudioLM. 
This result corresponds to the S2ST model presented in Table~\ref{table:top-level-results} trained on the S2ST mixture described in~\ref{subsec:method-mixtures}.

\paragraph{Discussion}
These observations are consistent with those reported in \cite{borsos2023soundstorm}, which found that compared to AudioLM, SoundStorm produces more intelligible speech when used to decode semantic audio tokens. This was measured by how faithfully the resulting audio matches a ground truth transcript when transcribed with an ASR system, which is similar to our setup.

\subsubsection{Impact of using PaLM-2
\label{subsec:exp-palm-2}}

In the following we explore the effect of using the PaLM-2 checkpoint vs the original PaLM model. 
PaLM-2 was trained with improved data and techniques compared to the original PaLM model, and was explicitly trained with parallel translation data.
We therefore aim to understand whether these improvements translate to gains in speech tasks.

\paragraph{Setup.}
We focus on speech-to-text tasks and do not consider S2ST.
We train two pairs of models on the largest datasets considered in Section~\ref{subsec:exp-scale-train-data}. 
For each dataset we train two models, one using the PaLM 8B checkpoint and the other using the PaLM-2 8B checkpoint.
Compared to the PaLM finetuning experiments, the optimization hyperparameters differed: we used a dropout rate of $0.2$ and a learning rate schedule of linear ramp-up to $10^{-4}$ followed by exponential decay to $10^{-5}$.

\paragraph{Results.}

See Table~\ref{table:palm-vs-palm-2}.
On the smaller mixture, we observe an improvement on the CoVoST2 AST task, and a minor degradation on VoxPopuli ASR and a more significant degradation on CoVoST2 ASR.
On the larger data mixture, we see that PaLM-2 exceeds PaLM on the Vox Populi ASR and CVSS AST tasks, and is slightly worse on CoVoST2 ASR. 
Our interpretation of these results is that the improved ability of PaLM-2 to perform text translation leads to an improvement for AST. The impact on ASR capabilities is mixed, where when using the full training mixture, PaLM 2 exhibits slightly worse ASR capabilities on CoVoST2 and slightly better ones on VoxPopuli ASR. 

\begin{table}
  \caption{Results for Experiment~\ref{subsec:exp-palm-2} showing impact of finetuning PaLM vs PaLM-2.}
  \label{table:palm-vs-palm-2}
  \centering
  \begin{tabular}{
    l
    l
    d{3.1}
    d{3.1}
    d{3.1}
   }
    \toprule
    \multirow{2}{*}{Datasets} & \multirow{2}{*}{Checkpoint}     
    & \multicolumn{1}{c}{VoxPopuli ASR}      &  \multicolumn{1}{c}{CoVoST2 ASR} &  \multicolumn{1}{c}{CVSS AST} \\
  & & \multicolumn{1}{c}{WER$\downarrow$} & \multicolumn{1}{c}{WER$\downarrow$} & \multicolumn{1}{c}{BLEU$\uparrow$} \\
    \midrule
    \multirow{2}{*}{Public + YT} 
      & PaLM    & $9.6$ & $13.8$ & $34.8$      \\
      &  PaLM-2 & $9.7$ & $17.4$ & $37.2$      \\
    \midrule
    \multirow{2}{*}{Public + YT + WMT/TED}
      &  PaLM   & $11.1$ & $15.1$ & $35.4$     \\
      &  PaLM-2 &  $9.8$ & $15.7$ & $37.8$     \\
    \bottomrule
  \end{tabular}
\end{table}

\paragraph{Discussion.}
While we do see a difference, we suspect that the different capabilities between PaLM and PaLM-2 are not as important in this setting as they might be for purely text-based tasks, since the addition of tokenized audio is novel for both models.

\subsubsection{Impact of architecture scale
\label{subsec:exp-scaling-architecture}}

In the following we investigate the impact of the model size on the downstream task performance. We use PaLM-2 for this and focus on the ASR and AST settings.

\paragraph{Setup.}

We train three PaLM-2 models of different sizes (128M, 1B, and 8B) using USM-v2 tokens with the same two largest datasets from Section~\ref{subsec:exp-scale-train-data} and observe their performance on our benchmark ASR and AST tasks.

\begin{table}
  \caption{Results for Experiment~\ref{subsec:exp-scaling-architecture} showing impact of architecture scale when using PaLM-2 checkpoints on AST/ASR tasks.}
  \label{table:scaling-architecture}
  \begin{adjustwidth}{-.5in}{-.5in}
  \centering
  \begin{tabular}{
    l
    c
    d{3.1}
    d{3.1}
    d{3.1}
    }
    \toprule
    \multirow{2}{*}{Datasets} & \multirow{2}{*}{Checkpoint size}     
      & \multicolumn{1}{c}{VoxPopuli ASR}      &  \multicolumn{1}{c}{CoVoST2 ASR} &  \multicolumn{1}{c}{CVSS AST} \\
    & & \multicolumn{1}{c}{WER$\downarrow$} & \multicolumn{1}{c}{WER$\downarrow$} & \multicolumn{1}{c}{BLEU$\uparrow$} \\
    \midrule
    \multirow{3}{*}{Public + YT} 
             & 128M & $15.9$ & $30.2$ & $16.6$      \\
             &   1B & $11.9$ & $21.5$ & $30.4$       \\
             &   8B &  $9.7$ & $17.4$ & $37.2$       \\
    \midrule
    \multirow{2}{*}{Public + YT + WMT/TED}  
             & 128M  & $16.4$ & $29.9$ & $18.3$     \\
             &  1B  &  $11.7$ & $17.3$ & $31.6$     \\
             &  8B  &   $9.8$ & $15.7$ & $37.8$     \\
    \bottomrule
  \end{tabular}
  \end{adjustwidth}
\end{table}

\paragraph{Results.} See Table~\ref{table:scaling-architecture}. We find that our results improve substantially with model size, with 42\% and 28\% reduction in WER for CVSS and VoxPopuli ASR tasks and over 13 points increase in BLEU scores for translation tasks respectively moving from 128M to 1B model on the full Public + YT + WMT/TED dataset. Increasing the model size further from 1B to 8B leads to additional gains of a further 10\% and 16\% reduction in WER for CVSS and VoxPopuli ASR tasks and a further 6.2 point improvement in BLEU score.
We find the scaling improvements also hold across different training datasets (e.g., Public + YT compared with Public + YT + WMT/TED).

\paragraph{Discussion.} As expected, performance on downstream ASR/AST tasks improves with larger model size. Our 1B sized model outperforms Whisper 1.5B Large model by over 5 BLEU points and 28\% reduction in WER for VoxPopuli ASR.

\section{Conclusion \label{sec:conclusion}}

We introduce \ours{}, a large language model that can process and generate speech and text interchangeably. \ours{} starts from a pre-trained text-based LLM and extends its vocabulary with discrete audio tokens. In doing so, the model can leverage its existing text capabilities while being finetuned to also consume and produce tokenized audio on a mixture of speech-text tasks. 
Moreover, by expressing the different tasks with textual tags, a single model can be trained on all tasks together. \ours{} demonstrates state-of-the-art results on speech translation benchmarks and competitive performance on speech recognition tasks, as well as zero-shot speech-to-text translation abilities on unseen language pairs. \ours{} also benefits from features of audio language models, such as voice prompting, and can perform S2ST with voice transfer of a superior quality compared to existing baselines, as measured by both automatic metrics and human raters.%

\paragraph{Limitations}
The fact that our model can natively produce audio is a consequence of the fact that we make use of tokenized audio.
This introduces a strong dependency on the quality of the audio tokenizer, as demonstrated in Section \ref{table:exp-tokens}.
We additionally empirically found it necessary to finetune the whole model, unlike a Flamingo-like \citep{alayrac2022flamingo} approach which freezes most of the weights and thus provides guarantees on preservation of the original capabilities of the model components.

\paragraph{Open questions}
There are numerous further avenues of research. 
One strand is around audio tokenization: what are desirable properties of audio tokens, how can we measure them, and how can we optimize for them?
Another is around evaluations. In comparison to text, the richness of the set of established benchmarks for generative text/audio tasks is less developed. This work has focused on speech recognition and speech translation, for which the benchmarks are more mature. The establishment of more benchmarks and metrics for generative audio tasks will help to accelerate research further.

\section*{Acknowledgements}
We would like to thank Nobuyuki Morioka and Yifan Ding for their help in re-creating the TTS-augmented WMT/TED dataset which was also used in~\cite{jia2022leveraging} and Adam Roberts and Ron Weiss for their advice and reviews. We would like to thank Slav Petrov, Colin Cherry and the PaLM-2 team for their advice and support.

\bibliography{references}

\newpage

\appendix

\section{Author Contributions}

Paul initiated the project, created the AudioPaLM model architecture and proved its viability with a number of pre-training, speech recognition and translation tasks, onboarded the team to the project, and contributed significantly to the write-up of this report. Chulayuth performed many audio tokenization experiments and completed the audio integration with PaLM2. Duc carried out the experiments around combined tasks (performing both ASR and AST in the same task), synthesized the WMT-derived speech-to-speech datasets and further developed the audio vocabulary. All of the above contributed to the best-performing configuration for speech-to-speech translation. Paul and Christian coordinated the write-up of this report. Duc and Chulayuth ran a majority of the ablation experiments.

Ankur, Johan, Tara and Yu collaborated on the research and development of USM-v2 tokens which led to our best-performing configuration across tasks. Jiahui and Zhishuai developed the token learning approach for images and Jiahui advised on the development of USM-v2 tokens. James, Wei and Yongqiang developed the large-scale tokenization and transcription infrastructure for USM models. Yongqiang, Wei and Félix curated the semi-supervised ASR datasets. Alexandru put in place data processing pipelines, improved our best mixture by adding a variety of ASR datasets, on AudioLM speech generation models, and initially worked on the USM-v2 audio tokens together with Johan. 

Danny worked on speech-to-speech translation and the ASR-BLEU metric for S2ST models together with Alexandru and Duc. Peter and Vicky worked on the PaLM2 integration and cascaded model baselines. Dalia significantly improved the best configuration of this report by adding TTS tasks and text-to-text and synthetic speech-to-speech datasets to the model’s task mixtures.

Eugene, Damien, Mihajlo, and Neil worked on AudioLM speech quality and in particular on making the translated voice consistent with the source voice and providing objective metrics. Mihajlo coordinated this effort, trained speech generation models, and created the website together with Hannah. Hannah further tuned the best models for the paper, analysed the zero-shot capabilities of the models, managed the rating tasks for subjective speech quality metrics, and performed a detailed training data analysis.

 Neil contributed significantly to the writing of this report. Marco identified the opportunity to leverage AudioLM for speech-to-speech translation and Zalán trained the very first such model. Zalán, Neil, and Marco provided guidance around AudioLM details and project planning. Michelle provided guidance on speech-to-speech baselines, Translatotron, and other related work. Lukas, Dirk, Matt and Johan supported and advised the team throughout the project. Christian initiated the project, coordinated the effort, and contributed with core ideas and technical work.

\newpage

\section{Detailed results of AST models performance}
\begin{table}[ht]
 \small
  \caption{BLEU scores on CoVoST2.}
  \label{table:covost-ast-bleu-per-language}
  \centering
  \setlength\tabcolsep{2pt}
  \noindent\makebox[1\textwidth]{
  \begin{tabular}{l c c c c c c c c c c c c c c c c c c c c c c }
    \toprule
    Model & \rotatebox{90}{Arabic (ar)} & \rotatebox{90}{Catalan (ca)} & \rotatebox{90}{Welsh (cy)} & \rotatebox{90}{German (de)} & \rotatebox{90}{Spanish (es)} &  \rotatebox{90}{Estonian (et)} & \rotatebox{90}{Persian (fa)} & \rotatebox{90}{French (fr)} & \rotatebox{90}{Indonesian (id)} & \rotatebox{90}{Italian (it)} & \rotatebox{90}{Japanese (ja)} & \rotatebox{90}{Latvian (lv)} & \rotatebox{90}{Mongolian (mn)} & \rotatebox{90}{Dutch (nl)} & \rotatebox{90}{Portuguese (pt)} & \rotatebox{90}{Russian (ru)} & \rotatebox{90}{Slovenian (sl)}  & \rotatebox{90}{Swedish (sv)}  & \rotatebox{90}{ Tamil (ta)}  & \rotatebox{90}{Turkish (tr)}  & \rotatebox{90}{Chinese (zh)} & \rotatebox{90}{All}\\
    \midrule
    Whisper 1.5B  \citep{radford2022whisper} & 39.7 & 31.8 & 21.5 & 36.3 & 40.1 & 15.0 & 19.3 & 36.4 & 48.1 & 30.9 & 26.1 & 13.9 & 0.1 & 41.2 & 51.6 & 43.3 & 21.6 & 42.9 & 4.2 & 28.3 & 18.0 & 29.1\\
    mSLAM-CTC 2B \citep{bapna2022mslam} & 19.3 & 35.4 & 6.7 & 35.9 & 41.0 & 22.6 & 9.7 & 39.0 & 8.8 & 37.3 & 3.3 & 26.8 & 0.8 & 37.6 & 42.8 & 48.4 & 32.3 & 38.5 & 0.6 & 24.2 & 10.0 & 25.2\\
    AudioPaLM 8B AST & 45.1 & 37.9 & 15.5 & 42.4 & 44.9 & 23.7 & 25.5 & 44.1 & 52.0 & 43.6 & 21.4 & 28.1 & 4.3 & 45.5 & 56.5 & 52.8 & 39.3 & 53.0 & 4.2 & 38.9 & 23.7 & 35.4\\
    AudioPaLM 8B S2ST & 45.5 & 36.4 & 19.4 & 41.4 & 43.4 & 27.2 & 28.4 & 43.2 & 54.3 & 42.9 & 24.4 & 33.3 & 5.8 & 43.4 & 55.5 & 54.3 & 41.8 & 53.8 & 6.9 & 37.5 & 21.4 & 36.2\\
    AudioPaLM-2 8B AST & 48.7 & 38.4 & 13.7 & 43.4 & 44.2 & 30.0 & 29.4 & 44.8 & 56.2 & 44.3 & 25.9 & 35.0 & 7.6 & 48.3 & 57.3 & 55.6 & 42.6 & 53.3 & 9.0 & 41.0 & 25.5 & 37.8\\
    \bottomrule
  \end{tabular}}
\end{table}

\begin{table}[ht]
 \small
  \caption{WER (\%) on Vox Populi.}
  \label{table:vp-wer-per-language}
  \centering
  \setlength\tabcolsep{3pt}
  \noindent\makebox[1\textwidth]{
  \begin{tabular}{l c c c c c c c c c c c c c c c c c c c c c c }
    \toprule
    Model & \rotatebox{90}{Czech (cs)} & \rotatebox{90}{German (de)} & \rotatebox{90}{English (en)} & \rotatebox{90}{Spanish (es)} &  \rotatebox{90}{Estonian (et)} & \rotatebox{90}{Finnish (fi)} & \rotatebox{90}{French (fr)} & \rotatebox{90}{Croatian (hr)} & \rotatebox{90}{Hungarian (hu)} & \rotatebox{90}{Italian (it)} & \rotatebox{90}{Lithuanian (lt)} & \rotatebox{90}{Dutch (nl)} & \rotatebox{90}{Polish (pl)} & \rotatebox{90}{Romanian (ro)} & \rotatebox{90}{Slovak (sk)} & \rotatebox{90}{Slovenian (sl)} & \rotatebox{90}{All} \\
    \midrule
    Whisper 1.5B \citep{radford2022whisper} & 12.6 & 11.2 & 7.0 & 18.6 & 28.7 & 12.4 & 11.4 & 16.1 & 13.8 & 19.0 & 33.2 & 12.9 & 7.8 & 14.4 & 15.4 & 27.9 & 13.6\\
    mSLAM-CTC 2B \citep{bapna2022mslam} & 6.8 & 8.7 & 7.0 & 8.4 & - & 8.7 & 9.4 & 9.1 & 8.4 & 15.4 & - & 10.5 & 6.4 & 7.8 & 6.0 & 15.1 & 9.1\\
    MAESTRO 600M \citep{chen2022maestro} & 6.9 & 7.9 & 6.3 & 7.2 & - & 8.6 & 7.9 & 8.5 & 7.1 & 13.3 & - & 9.2 & 5.7 & 7.3 & 5.2 & 14.2 & 8.1\\
    AudioPaLM 8B AST & 10.1 & 8.9 & 6.1 & 6.4 & - & 12.0 & 8.3 & 11.4 & 12.4 & 15.0 & - & 10.1 & 8.3 & 13.2 & 10.0 & 23.6 & 11.1\\
    AudioPaLM 8B S2ST & 13.6 & 10.6 & 6.5 & 7.6 & - & 17.6 & 10.0 & 17.4 & 9.8 & 17.0 & - & 11.1 & 7.6 & 22.7 & 8.6 & 64.3 & 16.0\\
    AudioPaLM-2 8B AST & 8.4 & 9.4 & 6.2 & 6.4 & - & 12.3 & 8.3 & 10.7 & 9.5 & 14.8 & - & 10.5 & 7.0 & 9.6 & 5.9 & 17.8 & 9.8\\
    \bottomrule
  \end{tabular}}
\end{table}

\section{Detailed results of S2ST models performance}
\begin{table}[ht]
 \small
  \caption{S2ST performance on CVSS, ASR-BLEU scores.}
  \label{table:cvss-s2st-bleu-per-language}
  \centering
  \setlength\tabcolsep{3pt}
  \noindent\makebox[1\textwidth]{
  \begin{tabular}{l c c c c c c c c c c c c c c c c c c c c c c }
    \toprule
    Model & \rotatebox{90}{Arabic (ar)} & \rotatebox{90}{Catalan (ca)} & \rotatebox{90}{Welsh (cy)} & \rotatebox{90}{German (de)} & \rotatebox{90}{Spanish (es)} &  \rotatebox{90}{Estonian (et)} & \rotatebox{90}{Persian (fa)} & \rotatebox{90}{French (fr)} & \rotatebox{90}{Indonesian (id)} & \rotatebox{90}{Italian (it)} & \rotatebox{90}{Japanese (ja)} & \rotatebox{90}{Latvian (lv)} & \rotatebox{90}{Mongolian (mn)} & \rotatebox{90}{Dutch (nl)} & \rotatebox{90}{Portuguese (pt)} & \rotatebox{90}{Russian (ru)} & \rotatebox{90}{Slovenian (sl)}  & \rotatebox{90}{Swedish (sv)}  & \rotatebox{90}{ Tamil (ta)}  & \rotatebox{90}{Turkish (tr)}  & \rotatebox{90}{Chinese (zh)} & \rotatebox{90}{All} \\
    \midrule
    Translatotron 2 \citep{jia2022leveraging} & 30.2 & 31.9 & 5.4 & 33.6 & 38.5 & 21.0 & 11.6 & 36.5 & 32.8 & 35.7 & 8.5 & 22.7 & 2.5 & 34.1 & 41.1 & 45.6 & 25.8 & 36.6 & 2.2 & 28.7 & 13.1 & 25.6 \\
    AudioPaLM 8B S2ST & 41.5 & 33.7 & 18.4 & 37.2 & 40.4 & 23.6 & 24.6 & 38.3 & 47.9 & 39.4 & 20.9 & 25.3 & 4.8 & 40.4 & 50.6 & 51.3 & 38.5 & 43.6 & 7.2 & 35.1 & 20.0 & 32.5\\
    \bottomrule
  \end{tabular}}
\end{table}

\newpage

\section{Detailed results of AST zero-shot performance
\label{appendix:ast-zero-shot}}

\begin{table}[ht]
 \small
  \caption{Zero-shot AST performance on FLEURS. BLEU scores for each language together with the number of hours of audio the model has been trained on in each language. The hours of training data for AudioPaLM do not include audio data seen in self-supervised training of audio tokenization models. The languages indicated with {\S} and {\textdagger} belong respectively to the ``AST observed'' and ``ASR observed'' sets used in Section~\ref{subsec:zeroshot_results}.}
  \label{table:fleurs-bleu-per-language}
  \centering
  \setlength\tabcolsep{3pt}
  \noindent\makebox[1\textwidth]{
  \begin{tabular}{l c c c c c c c c c c c c c c c c c c c c c }

    Model & \rotatebox{90}{Afrikaans\textsuperscript{\textdagger} (af)} & \rotatebox{90}{Amharic (am)} & \rotatebox{90}{Arabic\textsuperscript{\S} (ar)}  & \rotatebox{90}{Assamese (as)} & \rotatebox{90}{Asturian (ast)} & \rotatebox{90}{Azerbaijani\textsuperscript{\textdagger} (az)} & \rotatebox{90}{Belarusian\textsuperscript{\textdagger} (be)} & \rotatebox{90}{Bulgarian\textsuperscript{\textdagger} (bg)} & \rotatebox{90}{Bengali\textsuperscript{\textdagger} (bn)} & \rotatebox{90}{Bosnian (bs)} & \rotatebox{90}{Catalan\textsuperscript{\S} (ca)} & \rotatebox{90}{Cebuano (ceb)} & \rotatebox{90}{Kurdish (ckb)} & \rotatebox{90}{Chinese\textsuperscript{\S} (cmn)} & \rotatebox{90}{Czech\textsuperscript{\S} (cs)} & \rotatebox{90}{Welsh\textsuperscript{\S} (cy)} & \rotatebox{90}{Danish\textsuperscript{\textdagger} (da)} & \rotatebox{90}{German\textsuperscript{\S} (de)} & \rotatebox{90}{Greek\textsuperscript{\textdagger} (el)} & \rotatebox{90}{English (en)} & \rotatebox{90}{Spanish\textsuperscript{\S} (es)}\\ %
    \midrule
    Whisper 1.5B & 34.1 & 1.9 & 25.5 & 5.4 & - & 13.7 & 11.7 & 28.5 & 13.2 & 29.7 & 34.2 & - & - & 18.4 & 27.8 & 13.0 & 32.7 & 34.6 & 23.7 & 80.2 & 23.3\\
    \hspace{0.3cm}\small{AST training data (hours)}  & 330 & 32 & 2286 & 136 & 0 & 86 & 133 & 202 & 1988 & 219 & 236 & 0 & 0 & 11731 & 401 & 8263 & 386 & 4309 & 968 & 0 & 6693\\
    \hspace{0.3cm}\small{ASR training data (hours)}  & 4.1 & 0 & 739 & 0 & 0 & 47 & 2.4 & 86 & 1.3 & 11 & 1883 & 0 & 0 & 23446 & 192 & 73 & 473 & 13344 & 529 & 438218 & 11100\\
    \midrule
    AudioPaLM-2 8B & 34.7 & 3.8 & 29.0 & 9.3 & 30.8 & 16.2 & 15.1 & 35.5 & 15.9 & 35.7 & 42.5 & 10.3 & 4.0 & 21.3 & 34.5 & 7.2 & 37.9 & 38.7 & 18.8 & 77.2 & 26.9\\
    \hspace{0.3cm}\small{AST training data (hours)}  & 0 & 0 & 2 & 0 & 0 & 0 & 0 & 0 & 0 & 0 & 135 & 0 & 0 & 10 & 97 & 2 & 0 & 838 & 0 & 0 & 1887 \\
    \hspace{0.3cm}\small{ASR training data (hours)}  & 96 & 0 & 146 & 0.4 & 0 & 99 & 135 & 168 & 121 & 0 & 515 & 0 & 2 & 120 & 237 & 5 & 115 & 848 & 151 & 1465 & 2047\\
    \midrule
  \end{tabular}}
  \vspace{5pt}
  
  \setlength\tabcolsep{3pt}
  \noindent\makebox[1\textwidth]{
  \begin{tabular}{l c c c c c c c c c c c c c c c c c c c c c}

    Model & \rotatebox{90}{Estonian\textsuperscript{\S} (et)} & \rotatebox{90}{Persian\textsuperscript{\S} (fa)} & \rotatebox{90}{Fula (ff)} & \rotatebox{90}{Finnish\textsuperscript{\S} (fi)} & \rotatebox{90}{Filipino (fil)} & \rotatebox{90}{French\textsuperscript{\S} (fr)} & \rotatebox{90}{Irish (ga)} & \rotatebox{90}{Galician\textsuperscript{\textdagger} (gl)} & \rotatebox{90}{Gujarati\textsuperscript{\textdagger} (gu)} & \rotatebox{90}{Hausa (ha)} & \rotatebox{90}{Hebrew (he)} & \rotatebox{90}{Hindi\textsuperscript{\textdagger} (hi)} & \rotatebox{90}{Croatian\textsuperscript{\S} (hr)} & \rotatebox{90}{Hungarian\textsuperscript{\S} (hu)} & \rotatebox{90}{Armenian\textsuperscript{\textdagger} (hy)} & \rotatebox{90}{Indonesian\textsuperscript{\S} (id)} & \rotatebox{90}{Igbo (ig)} & \rotatebox{90}{Icelandic\textsuperscript{\textdagger} (is)} & \rotatebox{90}{Italian\textsuperscript{\S} (it)}  & \rotatebox{90}{Japanese\textsuperscript{\S} (ja)} & \rotatebox{90}{Javanese (jv)}\\
    \midrule
    Whisper 1.5B & 18.7 & 19.6 & - & 22.1 & 24.4 & 32.2 & - & 27.9 & 16.2 & 0.4 & 21.8 & 22.0 & 27.0 & 21.2 & 16.0 & 29.1 & - & 9.1 & 23.6 & 18.9 & 6.2 \\
    \hspace{0.3cm}\small{AST training data (hours)}  & 79 & 392 & 0 & 750 & 894 & 4481 & 0 & 368 & 208 & 8 & 418 & 5438 & 239 & 554 & 116 & 1174 & 0 & 84 & 2145 & 8860 & 622\\
    \hspace{0.3cm}\small{ASR training data (hours)}  & 41 & 24 & 0 & 1066 & 75 & 9752 & 0 & 9 & 0.3 & 0 & 688 & 12 & 91 & 379 & 13 & 1014 & 0 & 16 & 2585 & 7054 & 0\\
    \midrule
    AudioPaLM-2 8B & 31.7 & 25.7 & 0.29 & 29.3 & 15.6 & 36.5 & 0.3 & 34.7 & 12.2 & 0.6 & 0.4 & 21.7 & 30.6 & 29.2 & 10.2 & 34.2 & 0.3 & 17.8 & 27.8 & 11.1 & 9.7\\
    \hspace{0.3cm}\small{AST training data (hours)}  & 3 & 49 & 0 & 39 & 0 & 800 & 0 & 0 & 0 & 0 & 0 & 0 & 75 & 78 & 0 & 1 & 0 & 0 & 255 & 1 & 0 \\
    \hspace{0.3cm}\small{ASR training data (hours)}  & 163 & 165 & 0 & 175 & 0 & 857 & 0.2 & 1 & 123 & 0.7 & 0 & 101 & 34 & 239 & 126 & 121 & 0 & 91 & 338 & 181 & 0 \\
    \midrule
  \end{tabular}}
  \vspace{5pt}
  
  \setlength\tabcolsep{3pt}
  \noindent\makebox[1\textwidth]{
  \begin{tabular}{l c c c c c c c c c c c c c c c c c c c c c c }

    Model & \rotatebox{90}{Georgian\textsuperscript{\textdagger} (ka)} & \rotatebox{90}{Kamba (kam)} & \rotatebox{90}{Kabuverdianu (kea)} & \rotatebox{90}{Kazakh (kk)} & \rotatebox{90}{Khmer (km)} & \rotatebox{90}{Kannada (kn)} & \rotatebox{90}{Korean\textsuperscript{\textdagger} (ko)} & \rotatebox{90}{Kyrgyz (ky)} & \rotatebox{90}{Luxembourgish (lb)} & \rotatebox{90}{Ganda (lg)} & \rotatebox{90}{Lingala (ln)} & \rotatebox{90}{Lao (lo)} & \rotatebox{90}{Lithuanian\textsuperscript{\S} (lt)} & \rotatebox{90}{Luo (luo)} & \rotatebox{90}{Latvian\textsuperscript{\S} (lv)} & \rotatebox{90}{Maori (mi)} & \rotatebox{90}{Macedonian\textsuperscript{\textdagger} (mk)} & \rotatebox{90}{Malayalam\textsuperscript{\textdagger} (ml)} & \rotatebox{90}{Mongolian\textsuperscript{\S} (mn)} & \rotatebox{90}{Marathi\textsuperscript{\textdagger} (mr)} \\
    \midrule
    Whisper 1.5B (BLEU) & 2.4 & - & - & 5.4 & 6.1 & 11.6 & 21.3 & - & 16.8 & - & 1.0 & 11.0 & 14.0 & - & 14.3 & 10.2 & 27.7 & 16.7 & 1.0 & 12.9 \\
    \hspace{0.3cm}\small{AST training data (hours)} & 40 & 0 & 0 & 31 & 672 & 90 & 19938 & 0 & 10 & 0 & 20 & 20 & 99 & 0 & 68 & 1381 & 30 & 892 & 79 & 288\\
    \hspace{0.3cm}\small{ASR training data (hours)} & 0.6 & 0 & 0  & 12 & 1 & 4 & 7993 & 0 & 0 & 0 & 0 & 0.1 & 67 & 0 & 65 & 0 & 16 & 0.5 & 0 & 0.6\\
    \midrule
    AudioPaLM-2 8B (BLEU) & 13.6 & 1.6 & 29.4 & 9.5 & 0.1 & 4.8 & 19.4 & 8.61 & 16.1 & 1.6 & 0.7 & 9.5 & 26.8 & 0.6 & 30.5 & 1.2 & 30.8 & 12.2 & 10.1 & 17.1 \\
    \hspace{0.3cm}AST training data (hours) & 0 & 0 & 0 & 0 & 0 & 0 & 0 & 0 & 0 & 0 & 0 & 0 & 2 & 0 & 2 & 0 & 0 & 0 & 3 & 0\\
    \hspace{0.3cm}ASR training data (hours) & 137 & 0 & 0 & 0.2 & 0 & 0 & 157 & 0.7 & 0 & 26 & 0 & 0 & 172 & 0 & 150 & 0 & 104  & 110 & 4 & 90\\
    \midrule
  \end{tabular}}
  
  \end{table}
  
\begin{table}[t]
    \setcounter{table}{16}
    \caption{(continued) Zero-shot AST performance on FLEURS. BLEU scores for each language together with the number of hours of audio the model has been trained on in each language. The hours of training data for AudioPaLM do not include audio data seen in self-supervised training of audio tokenization models. The languages indicated with {\S} and {\textdagger} belong respectively to the ``AST observed'' and ``ASR observed'' sets used in Section~\ref{subsec:zeroshot_results}. In the last column ``All (82 languages)'', the average BLEU score and total number of AST/ASR training hours were computed over the 82 languages (out of 102) that were used to evaluate the Whisper model.}
 \small
  \label{table:fleurs-bleu-per-language-continued}
  \centering
  
  \setlength\tabcolsep{3pt}
  \noindent\makebox[1\textwidth]{
  \begin{tabular}{l c c c c c c c c c c c c c c c c c c c c c c }
    Model & \rotatebox{90}{Malay\textsuperscript{\textdagger} (ms)} & \rotatebox{90}{Maltese (mt)} & \rotatebox{90}{Myanmar (my)} & \rotatebox{90}{Norwegian (nb)} & \rotatebox{90}{Nepali\textsuperscript{\textdagger} (ne)} & \rotatebox{90}{Dutch\textsuperscript{\S} (nl)} &  \rotatebox{90}{Northern-Sotho (nso)} & \rotatebox{90}{Nyanja (ny)} & \rotatebox{90}{Occitan (oc)} & \rotatebox{90}{Oromo (om)} & \rotatebox{90}{Oriya (or)} & \rotatebox{90}{Punjabi (pa)} & \rotatebox{90}{Polish\textsuperscript{\S} (pl)} & \rotatebox{90}{Pashto (ps)} & \rotatebox{90}{Portuguese\textsuperscript{\S} (pt)} & \rotatebox{90}{Romanian\textsuperscript{\S} (ro)} & \rotatebox{90}{Russian\textsuperscript{\S} (ru)} & \rotatebox{90}{Sindhi (sd)} & \rotatebox{90}{Slovak\textsuperscript{\S} (sk)} & \rotatebox{90}{Slovenian\textsuperscript{\S} (sl)} \\
    \midrule
    Whisper 1.5B & 27.3 & 13.5 & 0.4 & 31.4 & 16.1 & 24.0 & - & - & 20.2 & - & - & 15.7 & 22.3 & 3.4 & 38.1 & 31.5 & 27.8 & 5.7 & 26.1 & 17.0 \\
    \hspace{0.3cm}\small{AST training data (hours)}  & 1691 & 41 & 59 & 322 & 133 & 1767 & 0 & 0 & 49 & 0 & 0 & 117 & 2200 & 63 & 3620 & 555 & 7687 & 46 & 144 & 395\\
    \hspace{0.3cm}\small{ASR training data (hours)} & 382 & 1 & 0.1 & 266 & 0.6 & 2077 & 0 & 0 & 0 & 0 & 0 & 0.8 & 4278 & 0 & 8573 & 356 & 9761 & 0 & 90 & 41\\
    \midrule
    AudioPaLM-2 8B & 31.9 & 12.4 & 0.0 & 34.6 & 16.2 & 29.1 & 1.1 & 1.4 & 22.9 & 0.3 & 8.9 & 6.0 & 25.3 & 0.4 & 38.4 & 35.7 & 31.2 & 1.4 & 32.3 & 27.4 \\
    \hspace{0.3cm}\small{AST training data (hours)}  & 0 & 0 & 0 & 0 & 0 & 96 & 0 & 0 & 0 & 0 & 0 & 0 & 174 & 0 & 10 & 156 & 18 & 0 & 52 & 17\\
    \hspace{0.3cm}\small{ASR training data (hours)}  & 126 & 0.7 & 0 & 0 & 170 & 195 & 0 & 0 & 0 & 0 & 0.2 & 0.3 & 267 & 0 & 16 & 246 & 179 & 0 & 191 & 170\\
    \midrule
  \end{tabular}}
  \vspace{5pt}
  
  \setlength\tabcolsep{3pt}
  \noindent\makebox[1\textwidth]{
  \begin{tabular}{l c c c c c c c c c c c c c c c c c c c c c}
    Model & \rotatebox{90}{Shona (sn)} & \rotatebox{90}{Somali (so)}  & \rotatebox{90}{Serbian\textsuperscript{\textdagger} (sr)} & \rotatebox{90}{Swedish\textsuperscript{\S} (sv)} & \rotatebox{90}{Swahili\textsuperscript{\textdagger} (sw)} & \rotatebox{90}{Tamil\textsuperscript{\S} (ta)} & \rotatebox{90}{Telugu\textsuperscript{\textdagger} (te)} & \rotatebox{90}{Tajik (tg)} & \rotatebox{90}{Thai\textsuperscript{\textdagger} (th)} & \rotatebox{90}{Turkish\textsuperscript{\S} (tr)} & \rotatebox{90}{Ukrainian\textsuperscript{\textdagger} (uk)} & \rotatebox{90}{Umbundu (umb)} & \rotatebox{90}{Urdu\textsuperscript{\textdagger} (ur)} & \rotatebox{90}{Uzbek (uz)} & \rotatebox{90}{Vietnamese\textsuperscript{\textdagger} (vi)} & \rotatebox{90}{Wolof (wo)} & \rotatebox{90}{Xhosa (xh)} & \rotatebox{90}{Yoruba (yo)} & \rotatebox{90}{Cantonese (yue)} & \rotatebox{90}{Zulu (zu)} & \rotatebox{90}{All (82 languages)}\\
    \midrule
    Whisper 1.5B & 1.8 & 0.7 & 32.5 & 35.3 & 7.2 & 9.2 & 12.5 & 14.5 & 16.1 & 26.6 & 29.4 & - & 17.2 & 6.0 & 20.4 & - & - & 1.4 & - & - & 17.9\\
    \hspace{0.3cm}\small{AST training data (hours)}  & 279 & 21 & 136 & 1055 & 282 & 1484 & 987 & 15 & 1635 & 2241 & 509 & 0 & 1990 & 4 & 1719 & 0 & 0 & 432 & 0 & 0 & 120.6k\\
    \hspace{0.3cm}\small{ASR training data (hours)}  & 0 & 0 & 28 & 2119 & 5 & 136 & 4 & 0.3 & 226 & 4333 & 697 & 0 & 104 & 0.3 & 691 & 0 & 0 & 0 & 0 & 0 & 117.1k\\
    \midrule
    AudioPaLM-2 8B & 0.4 & 0.9 & 34.3 & 40.4 & 9.1 & 15.0 & 13.3 & 17.1 & 15.0 & 30.1 & 26.9 & 0.9 & 13.3 & 17.2 & 15.6 & 0.3 & 0.2 & 0.7 & 7.4 & 1.9 & 20.4\\
    \hspace{0.3cm}\small{AST training data (hours)}  & 0 & 0 & 0 & 2 & 0 & 2 & 0 & 0 & 0 & 4 & 0 & 0 & 0 & 0 & 0 & 0 & 0 & 0 & 0 & 0 & 4.8k\\
    \hspace{0.3cm}\small{ASR training data (hours)}  & 0 & 0 & 122 & 129 & 126 & 125 & 109 & 0 & 156 & 145 & 138 & 0 & 116 & 0 & 149 & 0 & 0 & 0 & 1 & 0 & 12.8k\\
    \bottomrule
  \end{tabular}}
  \vspace{8cm}
\end{table}

\end{document}